\documentclass[]{c4ggroup}

\usepackage[utf8]{inputenc} % allow utf-8 input
\usepackage[T1]{fontenc}    % use 8-bit T1 fonts
\usepackage{hyperref}       
\usepackage{url}            % simple URL typesetting
\usepackage{booktabs}       % professional-quality tables
\usepackage{amsfonts}       % blackboard math symbols
\usepackage{nicefrac}       % compact symbols for 1/2, etc.
\usepackage{microtype}      % microtypography
\usepackage{xcolor}         % colors

\usepackage{amsmath}
\usepackage{amssymb}
\usepackage{graphicx}
\usepackage{float}
\usepackage{subcaption}
\usepackage{tabularx}
\usepackage{wrapfig}
\usepackage{multirow}
\usepackage{makecell}
\usepackage[most]{tcolorbox}
\usepackage{xcolor}
\usepackage[table]{xcolor}

\usepackage{ulem}

\definecolor{highlight}{gray}{0.90}
\definecolor{warmpaper}{HTML}{FAF8F3}
\definecolor{warmframe}{HTML}{E2DCCF}

\newtcolorbox{teaserbox}{
    enhanced,
    width=\linewidth,
    colback=warmpaper,
    colframe=warmframe,
    boxrule=0.3pt,
    arc=2mm,
    left=5pt,
    right=5pt,
    top=5pt,
    bottom=5pt,
    boxsep=0pt,
}

\title{MaskAlign: Token-Subset Representation Alignment for Efficient Diffusion Training}

\author[1,2]{Lianyu Pang\textsuperscript{*}}
\author[1,2,3]{Tianlin Pan\textsuperscript{*}}
\author[2]{Cheng Da}
\author[2]{Changqian Yu}
\author[2]{Huan Yang}
\author[2]{Kun Gai}
\author[1]{Song Guo}
\author[1]{Wenhan Luo\textsuperscript{\dag}}

\affiliation[1]{The Hong Kong University of Science and Technology}
\affiliation[2]{Kuaishou Technology}
\affiliation[3]{University of Chinese Academy of Sciences}

\contribution[*]{Equal contribution}
\contribution[\dag]{Corresponding author}

\begin{document}

\abstract{
Representation alignment with pretrained vision models has recently shown strong potential for accelerating diffusion transformer training. By aligning intermediate diffusion features with clean-image representations from self-supervised vision encoders, existing methods improve convergence and generation quality. However, such alignment also introduces a non-trivial constraint: diffusion models operate on noisy inputs whose usable information varies across timesteps, while the reference features are extracted from clean images. In this paper, we revisit this mismatch from a token-level perspective. We find that, under full-token representation alignment, tokens with large alignment-gradient norms exhibit a stable spatial preference, suggesting that the alignment objective does not affect all tokens uniformly and may encourage the model to rely on the complete set of clean-image tokens. To address this issue, we propose MaskAlign, a token-subset representation alignment method that applies alignment to randomly sampled token subsets during training. By exposing the model to different token subsets across iterations, MaskAlign reduces the dependence of representation alignment on the complete token set and encourages alignment behavior that is more stable under token-subset perturbations. To mitigate the information loss caused by directly dropping tokens, we further introduce a lightweight pre-mask token mixing block that shares information across tokens before masking. 
Experiments on ImageNet 256 $\times$ 256 show that MaskAlign consistently improves training convergence and generation quality. On SiT-XL/2, MaskAlign reaches the 8.3 FID level about 77$\times$ faster than vanilla SiT-XL/2 and the 5.9 FID level about 30$\times$ faster than SiT-XL/2 + REPA, measured by the number of training iterations required to reach the same FID level. It also reduces per-step training time by 11.6\% relative to REG, while improving FID from 3.4 to 2.8 at 400K iterations and from 2.7 to 2.4 at 1M iterations.
}

\maketitle

\begin{figure}[h]
    \centering

    \begin{teaserbox}
    \centering

    \newlength{\teaserimgsep}
    \setlength{\teaserimgsep}{2pt}

    % 左边：5+3 samples
    \begin{minipage}[t]{0.58\linewidth}
        \vspace{0pt}
        \centering
        \setlength{\tabcolsep}{0pt}
        \renewcommand{\arraystretch}{0}

        % 第一行：5 张图
        \begin{tabular}{@{}c@{\hspace{\teaserimgsep}}c@{\hspace{\teaserimgsep}}c@{\hspace{\teaserimgsep}}c@{\hspace{\teaserimgsep}}c@{}}
            \includegraphics[width=\dimexpr(\linewidth-4\teaserimgsep)/5\relax]{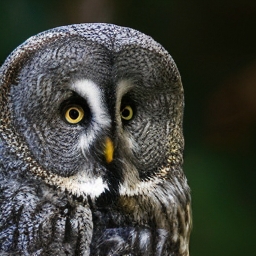} &
            \includegraphics[width=\dimexpr(\linewidth-4\teaserimgsep)/5\relax]{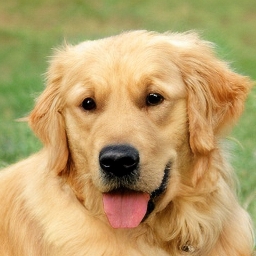} &
            \includegraphics[width=\dimexpr(\linewidth-4\teaserimgsep)/5\relax]{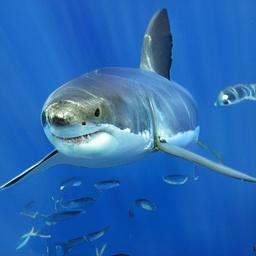} &
            \includegraphics[width=\dimexpr(\linewidth-4\teaserimgsep)/5\relax]{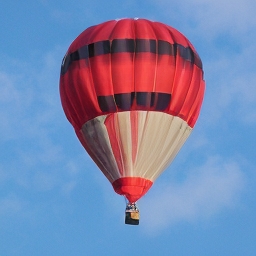} &
            \includegraphics[width=\dimexpr(\linewidth-4\teaserimgsep)/5\relax]{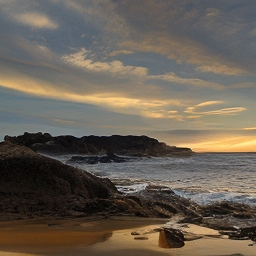}
        \end{tabular}

        \vspace{\teaserimgsep}

        % 第二行：3 张图
        \begin{tabular}{@{}c@{\hspace{\teaserimgsep}}c@{\hspace{\teaserimgsep}}c@{}}
            \includegraphics[width=\dimexpr(\linewidth-2\teaserimgsep)/3\relax]{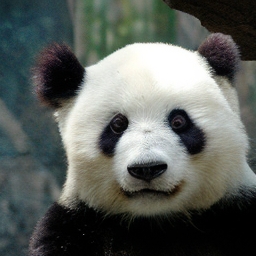} &
            \includegraphics[width=\dimexpr(\linewidth-2\teaserimgsep)/3\relax]{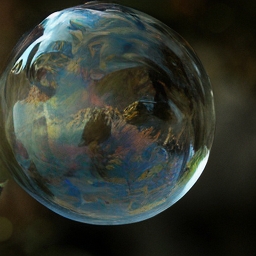} &
            \includegraphics[width=\dimexpr(\linewidth-2\teaserimgsep)/3\relax]{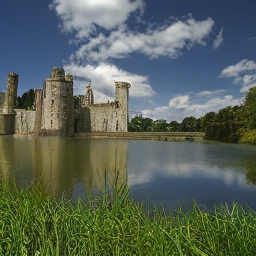}
        \end{tabular}
    \end{minipage}%
    \hfill
    % 右边：FID 曲线
    \begin{minipage}[t]{0.4\linewidth}
        \vspace{0pt}
        \centering
        \includegraphics[width=\linewidth]{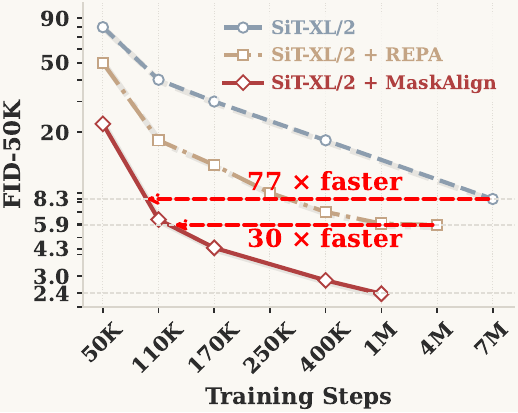}
    \end{minipage}

    \end{teaserbox}

    \caption{
    MaskAlign generates high-quality ImageNet $256 \times 256$ samples and reaches comparable FID with substantially fewer training iterations, showing faster convergence.
    }
    \label{fig:teaser}
\end{figure}
\section{Introduction}
Diffusion models have advanced significantly in recent years~\cite{ddpm,ddim,nichol2021glide,balaji2023ediffi,ldm,flux,imagen}. Latent diffusion models (LDMs)~\cite{ldm} utilize a Variational Autoencoder (VAE)~\cite{vae} to shift the image generation process from the pixel space to the latent space. 
DiT~\cite{dit} improves scalability through a transformer-based architecture, and SiT~\cite{sit} further enhances performance by employing continuous-time stochastic interpolants. 
Despite these advances, training high-quality image generation models at scale remains prohibitively expensive, requiring enormous computational resources and training time.

Recent studies have utilized pretrained self-supervised vision models to accelerate diffusion training, as their rich visual features can guide the generative model toward better representations.
REPA~\cite{repa} is a representative method in this direction, directly aligning intermediate diffusion features with those of a vision encoder to improve convergence and generation quality.
Following this paradigm, subsequent studies have improved representation-based diffusion training through class tokens~\cite{reg}, shared latent feature coupling~\cite{redi}, VAE-level representation alignment~\cite{leng2025repae}, and other alignment-based objectives~\cite{haste,reglue,singh2025matters}.

While these methods have proven highly successful at speeding up diffusion training, representation alignment introduces a non-trivial training constraint.
Pretrained vision models usually take clean images as input, so their features encode rich visual and semantic information.
In contrast, diffusion models operate on noisy inputs, where the usable information varies with the noise level and the model's intermediate features shift accordingly.
This leads to a potential mismatch: the diffusion model is encouraged to match tokens derived from a clean image, even though its own input is noisy and only partially informative. 

\begin{figure}[t]
    \centering
    \captionsetup[subfigure]{labelfont=bf}
    \begin{subfigure}{0.21\textwidth}
        \centering
        \includegraphics[width=\textwidth]{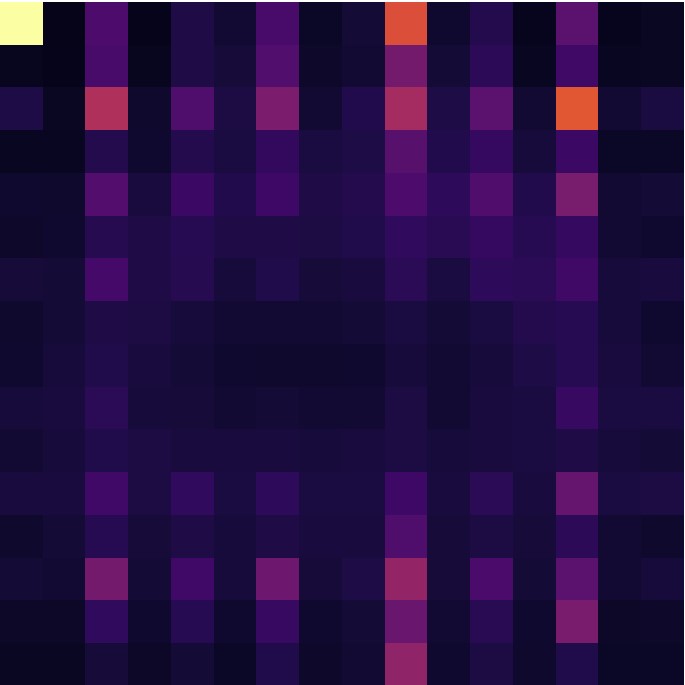} 
        \caption{Full-token heatmap}
        \label{fig:p_map_mask_0}
    \end{subfigure}
    \hfill
    \begin{subfigure}{0.21\textwidth}
        \centering
        \includegraphics[width=\textwidth]{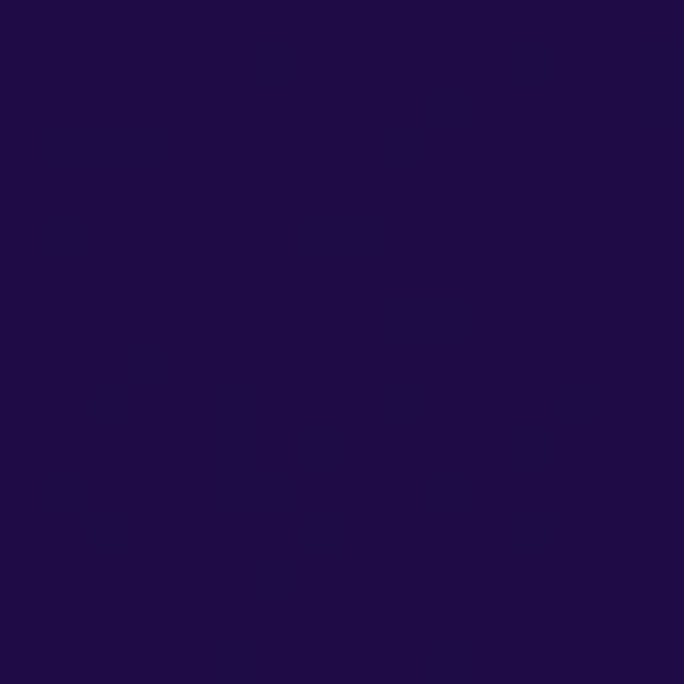}
        \caption{$25\%$ mask heatmap}
        \label{fig:p_map_mask_0.25}
    \end{subfigure}
    \hfill
    \begin{subfigure}{0.25\textwidth}
        \centering
        \includegraphics[width=\textwidth]{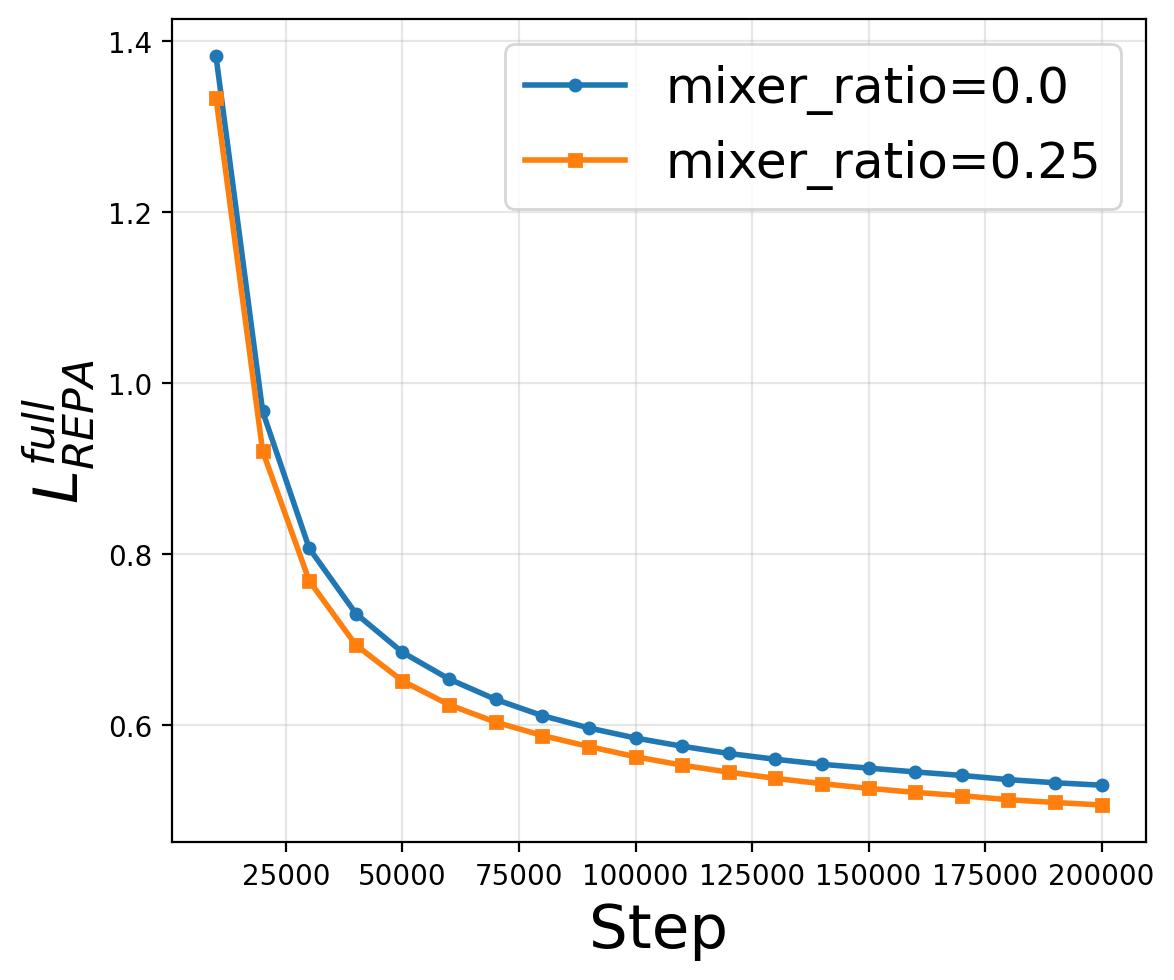}
        \caption{Alignment loss}
        \label{fig:L_repa}
    \end{subfigure}
    \hfill
    \begin{subfigure}{0.25\textwidth}
        \centering
        \includegraphics[width=\textwidth]{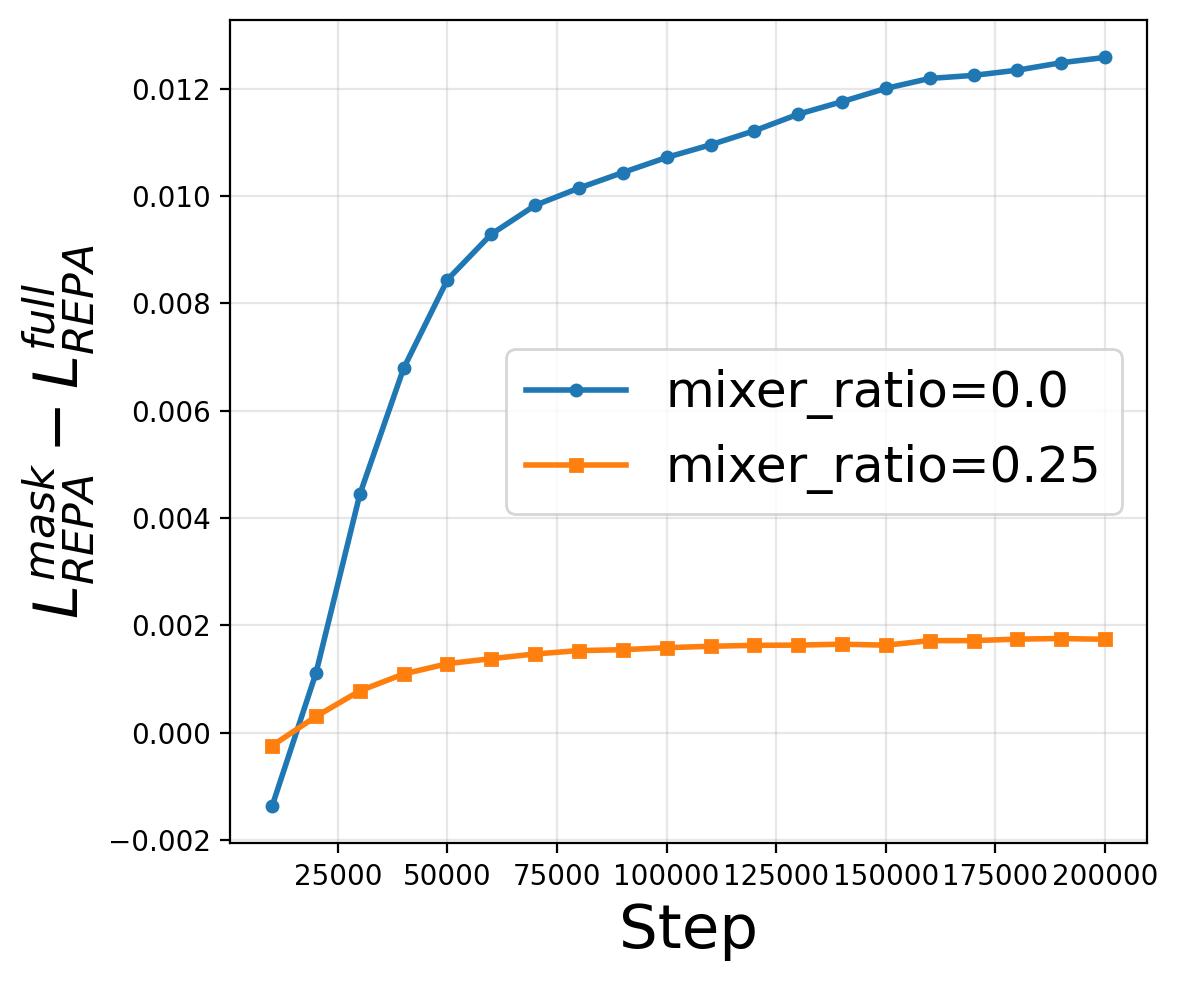}
        \caption{Alignment-loss gap}
        \label{fig:g_r}
    \end{subfigure}
    
    % \caption{
    % Token-level behavior and alignment stability under token masking.
    % \textbf{(a,b)} Heatmaps show the probability that each spatial position appears among the top-$10\%$ tokens ranked by alignment-gradient norm.
    % Both heatmaps use the same color range $[0,0.8]$, where brighter colors indicate a higher probability. 
    
    % \textbf{(a)} Under full-token alignment, high-gradient tokens show a stable spatial preference.
    % \textbf{(b)} Under a $25\%$ mask ratio, the concentrated spatial pattern is largely reduced.
    % \textbf{(c)} MaskAlign lowers the full-token alignment loss during training.
    % \textbf{(d)} MaskAlign narrows the alignment-loss gap $L^{\text{mask}}_{\text{REPA}}-L^{\text{full}}_{\text{REPA}}$ between randomly masked and full-token inputs, indicating more stable alignment behavior under token-subset perturbations.
    % }
    \caption{
        Token-level behavior and alignment stability under token masking.
        \textbf{(a,b)} Heatmaps show the probability that each spatial position appears among the top-$10\%$ alignment-gradient tokens, using the same color range $[0,0.8]$. For reference, a uniform distribution would correspond to approximately $10\%$ for each position.
        \textbf{(a)} Full-token alignment exhibits a stable spatial preference.
        \textbf{(b)} A $25\%$ mask ratio substantially reduces this concentrated pattern.
        \textbf{(c)} MaskAlign lowers the full-token alignment loss.
        \textbf{(d)} MaskAlign narrows the alignment-loss gap
        $L^{\mathrm{mask}}_{\mathrm{REPA}}-L^{\mathrm{full}}_{\mathrm{REPA}}$,
        indicating improved stability under token-subset perturbations.
    }
    \label{fig:observation}
\end{figure}

We inspect this mismatch at the token level by studying the gradient distribution of the alignment loss, as shown in Figure~\ref{fig:observation}.
Figure~\ref{fig:p_map_mask_0} shows that certain spatial positions are more likely to produce top-$10\%$ gradient-norm tokens than others, even after averaging over many images.
These high-gradient tokens form a stable spatial pattern, suggesting that the alignment objective does not affect all tokens uniformly.
Since the alignment loss is applied to all clean-image tokens unconditionally, it may encourage a feature-fitting shortcut that matches clean feature patterns without ensuring their usefulness under noisy denoising conditions.

Building on these observations, we adopt a dropout-like strategy inspired by random feature dropping for preventing co-adaptation~\cite{baldi2013understanding,wager2013dropout}: we randomly mask patch tokens during alignment to reduce shortcuts that rely on the complete token set.
By averaging the alignment objective over random token subsets, this strategy disrupts stable patterns of concentrated gradients and encourages alignment signals that remain effective across different subsets.
However, directly dropping tokens may disrupt fine-grained spatial patterns. We therefore add a lightweight pre-mask mixing block to share information across tokens before masking.

Figures~\ref{fig:L_repa} and~\ref{fig:g_r} show that masked training not only reduces the alignment loss, but also narrows the alignment-loss gap between randomly masked and full-token inputs. This indicates that the learned alignment behavior becomes less sensitive to token-subset perturbations. Figure~\ref{fig:teaser} further reports FID over training steps on ImageNet $256 \times 256$~\cite{deng2009imagenet}. MaskAlign reaches the same FID levels with substantially fewer training iterations: it reaches the 8.3 FID level about $77\times$ faster than vanilla SiT-XL/2 and the 5.9 FID level about $30\times$ faster than SiT-XL/2 + REPA. Here, speedup is measured by the number of training iterations required to reach the same FID level. Together with the lower per-step cost introduced by token masking, these results show that MaskAlign improves both convergence and training efficiency.

In summary, our contributions are as follows:
\begin{itemize}
\item We analyze the training behavior of representation alignment at the token level. We find that, under full-token representation alignment, gradients are non-uniformly distributed across patch tokens, with high-gradient tokens exhibiting a stable spatial preference.

\item We propose MaskAlign, a random token masking strategy that applies alignment to randomly sampled token subsets instead of the complete token set. Motivated by dropout's ability to prevent co-adaptation, MaskAlign discourages feature-fitting shortcuts and encourages alignment signals that remain stable across different token subsets. We further introduce a lightweight pre-mask token mixer to reduce the information loss caused by directly dropping tokens.

\item We validate the effectiveness of MaskAlign on ImageNet $256 \times 256$. MaskAlign reaches the same FID levels with substantially fewer training iterations, achieving about $77\times$ faster convergence than vanilla SiT-XL/2 at the 8.3 FID level and about $30\times$ faster convergence than SiT-XL/2 + REPA at the 5.9 FID level. It also reduces the full-token alignment loss and improves alignment stability under token-subset perturbations.
\end{itemize}

\section{Related Work}
\paragraph{Generative Models for Image Generation.} Early methods, such as DDPM~\cite{ddpm} and DDIM~\cite{ddim}, generate images by denoising directly in the pixel space. In contrast, Latent Diffusion Models (LDMs)~\cite{ldm} first use a VAE~\cite{vae} to map images into a latent space before performing the denoising process, which significantly improves both training and inference efficiency. Early LDMs~\cite{ldm,nichol2021glide,balaji2023ediffi} utilized U-Net as their foundational architecture. Later, the transformer-based DiT~\cite{dit} architecture was adopted to enhance scalability. Most recently, SiT~\cite{sit}, which incorporates continuous-time stochastic interpolants, has further improved the training efficiency of LDMs. Despite these significant advancements, training large-scale image generation models remains a challenge that requires substantial computational resources.

\paragraph{Efficient Training via Token Masking.}
Accelerating the training of LDMs has been a major research focus. Token masking provides a viable solution approach. Methods like MDT~\cite{gao2023mdtv2} and MaskDiT~\cite{maskdit} reduce the number of input tokens during training. By forcing the model to predict all tokens from a subset of tokens, these methods encourage the model to better learn the contextual relationships within the image. To mitigate the information loss caused by masking, MicroDiT~\cite{microdiffusion} first uses a lightweight mixer to aggregate token information before applying the mask. Furthermore, TREAD~\cite{krause2025tread} observed minimal output variations across intermediate DiT layers and proposed routing a portion of tokens to skip these layers, thereby avoiding the masking-induced information loss. Different from these methods, we do not use masking as a reconstruction task over missing tokens. Instead, we use random masking to construct token subsets for representation alignment, with the prediction and alignment losses computed on the preserved class token and visible patch tokens.

\paragraph{Representation Alignment with External Models.} Representation alignment has recently become an active research direction. REPA~\cite{repa} observed that DiT models also capture image semantics during training. It proposed aligning the intermediate features of DiTs with the output features of a strong pretrained vision model to improve both training efficiency and final generation performance. Building upon this, REG~\cite{reg} and ReDi~\cite{redi} improved the alignment strategy, enabling the model to better learn the semantic information from the pretrained vision model. REPA-E~\cite{leng2025repae} employs the REPA loss to train a VAE model, substantially enhancing the overall generation quality. However, HASTE~\cite{haste} identified a conflict between the two optimization objectives in REPA. Specifically, in the later stages of training, forcing the intermediate DiT features to align with the output features of an external pretrained vision model can degrade the model’s generation performance. To prevent this degradation, they introduced an early stopping mechanism. Different from these works, we study representation alignment from a token-level perspective and show that random token subsets can improve alignment stability during training.
\begin{figure}[t]
    \centering
    % 直接调用 example-image，可以随意设置宽高
    \includegraphics[width=\textwidth]{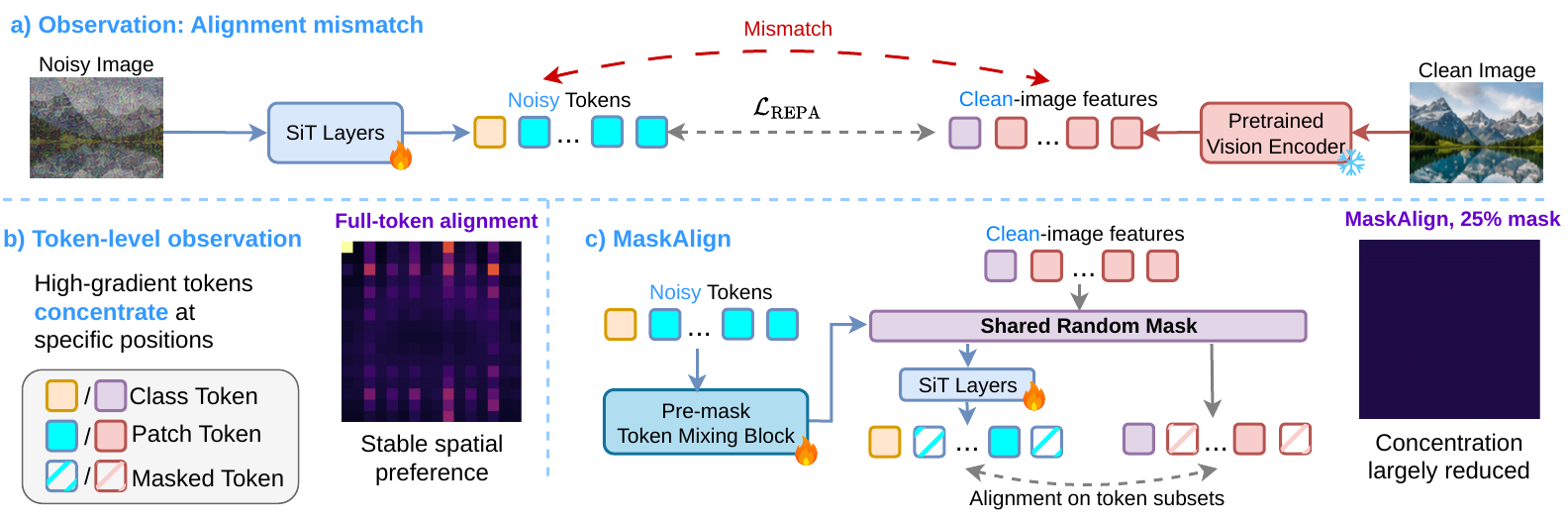} 
    \caption{
Overview of MaskAlign.
\textbf{a)} Representation alignment matches noisy diffusion tokens with clean-image features extracted by a pretrained vision encoder, leading to a potential mismatch across denoising timesteps.
\textbf{b)} Full-token alignment exhibits a stable spatial preference, where high-gradient tokens concentrate at specific spatial positions.
\textbf{c)} MaskAlign first applies pre-mask token mixing and then uses a shared random mask to compute representation alignment on token subsets while preserving the class token. With a $25\%$ mask ratio, MaskAlign substantially reduces the concentrated spatial pattern.
}
    \label{fig:framework}
\end{figure}

\section{Preliminaries}
\paragraph{Denoising Diffusion Probabilistic Models (DDPM).}
As a prominent family of generative models, diffusion models~\cite{ddpm,ddim,dit} synthesize high-fidelity images through a process of iterative denoising.  Under the common noise-prediction parameterization, the training objective minimizes the distance between the injected noise and the network prediction:
\begin{align}
  \mathcal{L}_{\text{diffusion}}=\mathbb{E}_{z, c, \varepsilon, t}\left[\left\|\varepsilon-\varepsilon_{\theta}\left(z_{t}, t, c\right)\right\|_{2}^{2}\right],
  \label{eq:diffusion_loss}
\end{align}
Here, the network $\varepsilon_{\theta}$ predicts the noise added to the corrupted input $z_t$, conditioned on the timestep $t$ and the context vector $c$.

\paragraph{Scalable Interpolant Transformers (SiT).}
\label{sec:sit}
Our method follows the SiT framework~\cite{sit}, which is derived from the stochastic interpolant formulation~\cite{lipman2022flow}. Let $ z_*$ denote a clean image, and let a pretrained VAE encoder $\mathcal{E}_z$ map it into the latent space as $z_0 \in \mathbb{R}^{D_z \times H_z \times W_z}$. Based on this latent representation, we construct a continuous-time interpolation process defined as:
\begin{align}
z_t = \alpha_t z_0 + \sigma_t \epsilon_z, \quad \epsilon_z \sim \mathcal{N}(0, I), \; t \in [0,1]
\label{eq:sit_def}
\end{align}
where the coefficients satisfy boundary conditions $\alpha_0 = \sigma_1 = 1 $ and $\alpha_1 = \sigma_0 = 0$. As $ t $ increases, $ \alpha_t $ decreases while $ \sigma_t $ increases accordingly.

The SiT model adopts a Transformer architecture composed of $K$ stacked blocks to learn a velocity function $ v_\theta(z_t, t) $. Training is carried out by minimizing the following velocity matching objective:
\begin{align}
    \mathcal{L}_{\mathrm{SiT}}
    =
    \mathbb{E}_{z,\epsilon_z,t}
    \left[
    \left\|
    v_\theta(z_t,t)
    - \dot{\alpha}_t z_0
    - \dot{\sigma}_t \epsilon_z
    \right\|_2^2
    \right].
    \label{eq:sit}
\end{align}
In our implementation, we use a linear parameterization $\alpha_t = 1 - t$ and $\sigma_t = t$, which results in constant time derivatives $\dot{\alpha}_t = -1$ and $\dot{\sigma}_t = 1$, unless stated otherwise.
\section{Token-level Analysis}
\subsection{Alignment-Gradient Distribution}
Representation alignment trains a diffusion model by matching its intermediate features with clean-image representations extracted by a pretrained vision encoder. 
However, the diffusion model operates on noisy inputs, where the usable information varies with the noise level. 
At different timesteps, the model may rely on different visual cues, from coarse structures under high noise levels to finer details under lower noise levels. 
In contrast, the reference features are always extracted from clean images. 
This creates a potential mismatch between the clean-image reference features and the model's noisy intermediate features. 
We therefore inspect this mismatch at the token level by analyzing the gradient distribution of the alignment loss.

We first consider the full-token alignment setting, where all patch tokens are aligned with their corresponding clean-image reference features. Since the class token has no spatial position, our token-level heatmap analysis focuses on patch tokens.
Given the hidden state $h_i^{[\ell_a]}$ at layer $\ell_a$ and the reference feature $r_i$, the alignment loss is defined as
\begin{align}
\mathcal L_{\mathrm{REPA}}
=
-\frac{1}{N}
\sum_{i=1}^{N}
\mathrm{sim}\left(
r_i,
h_\phi(h_i^{[\ell_a]})
\right),
\end{align}
where $h_\phi(\cdot)$ is the alignment projector. We omit the expectation over samples and timesteps for simplicity.

To examine how this objective affects training, we analyze the gradient norms of $\mathcal L_{\mathrm{REPA}}$ with respect to the hidden states at layer $\ell_a$. We focus on this layer because it is where the alignment supervision is explicitly injected through the projector.

Let $h_i^{[\ell_a]}$ denote the hidden state of the $i$-th patch token at the alignment layer $\ell_a$. We compute the alignment-gradient norm for each patch token as
\begin{align}
g_i^{\mathrm{align}} =
\left\|
\frac{\partial \mathcal L_{\mathrm{REPA}}}
{\partial h_i^{[\ell_a]}}
\right\|_2 .
\end{align}
For each image, we select the top-$k$ patch tokens with the largest gradient norms. We then compute the probability that each spatial position appears in this top-$k$ set across multiple images.

As shown in Figure~\ref{fig:p_map_mask_0}, certain spatial positions remain more likely to appear in the top-$k$ set, even after averaging over many images.
The largest spatial probability is about $21\times$ the smallest, suggesting that this preference cannot be explained by minor random fluctuations.
This indicates that the alignment-loss gradients are not uniformly distributed: tokens with large gradient norms tend to concentrate at certain spatial positions.
Therefore, we seek to reduce the dependence of representation alignment on the complete token set.

\subsection{Motivation for Token-Subset Alignment}
The token-level observation above suggests that full-token alignment may repeatedly reinforce high-gradient tokens at certain spatial positions. Since the reference features are extracted from clean images, the model may learn feature-fitting shortcuts that reduce the alignment loss for the complete token set but do not remain consistently useful under noisy denoising conditions.

Building on this observation, MaskAlign applies random token masking during representation alignment. Motivated by random feature dropping for preventing co-adaptation~\cite{baldi2013understanding,wager2013dropout}, we randomly sample patch-token subsets during training. As the visible token subsets vary across iterations, shortcuts that rely on the complete token set are less consistently reinforced. The model is therefore encouraged to rely on alignment signals that remain stable across different random token subsets.
\section{MaskAlign}
\subsection{Framework}
The overall framework of MaskAlign is shown in Figure~\ref{fig:framework}. 
Following REG, MaskAlign prepends a class token with global semantics to the patch tokens.
During training, the class token is always preserved, and representation alignment is applied to this token together with a randomly sampled subset of patch tokens.
Before masking, we apply lightweight pre-mask token mixing to share information across tokens and mitigate the disruption from dropping patch tokens. The mixed class token and visible patch tokens are then fed into the diffusion transformer.
Random token masking is used only during training; at inference time, all tokens are retained.

\paragraph{Pre-mask Token Mixing and Random Masking.}
Following Sec.~\ref{sec:sit}, let $z_*$ denote a clean image and let
$z_0=\mathcal{E}_z(z_*)\in \mathbb{R}^{D_z\times H_z\times W_z}$
be its clean latent. At timestep $t$, the noisy latent is constructed as
\begin{align}
    z_t = \alpha_t z_0 + \sigma_t \epsilon_z, 
    \quad
    \epsilon_z \sim \mathcal{N}(0,I).
\end{align}

The noisy latent $z_t$ is patchified and projected into a sequence of patch tokens
$x_t^0 = \{x_{t,1}^0,\ldots,x_{t,N}^0\}\in \mathbb{R}^{N\times D}$,
where $N$ is the number of patch tokens and $D$ is the hidden dimension.
Following REG, we prepend a class token $c_t^0$ to form
$H_t^0 = [c_t^0, x_t^0]\in\mathbb{R}^{(N+1)\times D}$.

Before random token masking, we apply a lightweight pre-mask token mixing block $M_\psi(\cdot)$ to share information across tokens:
\begin{align}
\bar H_t^0
=
[\bar c_t^0, \bar x_t^0]
=
M_\psi(H_t^0,t,y),
\end{align}
where $y$ denotes the class condition. This step mitigates the disruption caused by directly dropping patch tokens.

We then sample a binary keep mask $m\in\{0,1\}^{N}$ over patch tokens, where $m_i=1$ indicates that the $i$-th patch token is visible.
Let $S(m)=\{i\mid m_i=1\}$ denote the visible patch-token indices, with $N_m=|S(m)|$.
The class token is always preserved, while random masking is applied only to patch tokens:
\begin{align}
\widetilde H_t^0(m) = \big[ \bar c_t^0, \{\bar x_{t,i}^0\}_{i\in S(m)} \big] \in
\mathbb{R}^{(1+N_m)\times D},
\end{align}
where $[\,\cdot\,]$ denotes sequence concatenation. 
The masked sequence $\widetilde H_t^0(m)$ is then fed into the following SiT blocks. At layer $\ell$, the transformer produces
\begin{align}
H_t^{[\ell]}(m) = \big[ h_{t,\mathrm{cls}}^{[\ell]}(m), \{h_{t,i}^{[\ell]}(m)\}_{i\in S(m)} \big].
\end{align}

\paragraph{Training Losses.}
After random token masking, the prediction loss is computed on the preserved class token and the visible patch tokens.
Let $r^*=\{r_{\mathrm{cls}}, r_1,\ldots,r_N\}$ be the reference representation extracted from the clean image by the pretrained vision encoder, where $r_{\mathrm{cls}}$ denotes the projected clean class token.
Following REG, we construct the noisy class token as
$c^0_t=\alpha_t r_{\mathrm{cls}}+\sigma_t\epsilon_{\mathrm{cls}}$,
with target velocity
$v_{\mathrm{cls}}^*(t)=\dot{\alpha}_t r_{\mathrm{cls}}+\dot{\sigma}_t\epsilon_{\mathrm{cls}}$.
For each visible patch token $i\in S(m)$, let $\hat v_i(m,t)$ and
$v_i^*(t)=\dot{\alpha}_t z_{0,i}+\dot{\sigma}_t\epsilon_{z,i}$
denote the predicted and target velocities.
For the class token, let $\hat v_{\mathrm{cls}}(m,t)$ denote the predicted velocity.
We weight the class-token prediction loss by $\beta$:
\begin{align}
\mathcal{L}_{\mathrm{pred}}
=
\mathbb{E}_{z^*,\epsilon_z,\epsilon_{\mathrm{cls}},t,m}
\left[
\frac{1}{N_m}
\sum_{i\in S(m)}
\left\|
\hat v_i(m,t)-v_i^*(t)
\right\|_2^2
+
\beta
\left\|
\hat v_{\mathrm{cls}}(m,t)-v_{\mathrm{cls}}^*(t)
\right\|_2^2
\right].
\end{align}

At alignment layer $\ell_a$, we define the visible alignment index set as
$\mathcal{A}(m)=\{\mathrm{cls}\}\cup S(m)$, where the class token is always included.
The projector $h_\phi(\cdot)$ maps hidden states into the reference feature space.
For each $a\in\mathcal{A}(m)$, let $r_a$ and $h_{t,a}^{[\ell_a]}(m)$ denote the corresponding reference feature and hidden state. The alignment loss is then
\begin{align}
\mathcal{L}_{\mathrm{REPA}}
:=
-\mathbb{E}_{z_*,\epsilon_z,t,m}
\left[
\frac{1}{|\mathcal{A}(m)|}
\sum_{a\in\mathcal{A}(m)}
\mathrm{sim}\!\left(
r_a,
h_\phi(h_{t,a}^{[\ell_a]}(m))
\right)
\right],
\end{align}
where $|\mathcal{A}(m)|=N_m+1$. 
The final training objective is 
$\mathcal{L}_{\mathrm{total}}=\mathcal{L}_{\mathrm{pred}}+\lambda\mathcal{L}_{\mathrm{REPA}}$,
where $\lambda$ controls the strength of representation alignment.
% \paragraph{Training and inference.}

\subsection{Measuring Alignment Stability}
To assess alignment stability under token-subset perturbations, we compare the full-token and masked-input alignment losses.
Let $\mathcal{L}^{\mathrm{full}}_{\mathrm{REPA}}$ denote the alignment loss computed using the class token and all patch tokens, and let $\mathcal{L}^{\mathrm{mask}}_{\mathrm{REPA}}$ denote the alignment loss computed using the class token and a randomly sampled subset of patch tokens.  We define the alignment-loss gap as
\begin{align}
G_r =
\mathcal L_{\mathrm{REPA}}^{\mathrm{mask}}
-
\mathcal L_{\mathrm{REPA}}^{\mathrm{full}}.
\end{align}
A smaller $G_r$ indicates that the alignment loss is less sensitive to token-subset perturbations, and thus the learned alignment behavior is more stable across random token subsets. 

Figures~\ref{fig:L_repa} and~\ref{fig:g_r} report the full-token alignment loss and the alignment-loss gap $G_r$ for REG and MaskAlign under a $25\%$ mask ratio.
At 200K steps, the gap of MaskAlign is only $13.8\%$ of that of REG, showing that MaskAlign is much less sensitive to token-subset perturbations.
In contrast, the larger gap of REG suggests stronger dependence on the complete token set.
These results provide evidence that random token masking encourages more stable alignment behavior under token-subset perturbations.

\definecolor{highlight}{gray}{0.90}

\begin{table}[t]
\begin{minipage}{0.4\textwidth}
\centering
\caption{FID comparison during training on ImageNet 256 $\times$ 256 without CFG.}
\scriptsize
\label{tab:wio_cfg}
\renewcommand{\arraystretch}{1.063}
\begin{tabular}{lccc}
\toprule
Method & \#Params & Iter. & FID$\downarrow$ \\
\midrule
SiT-B/2 & 130M & 400K & 33.0 \\
\midrule
REPA & 130M & 400K & 24.4 \\
REG &  132M & 400K & 15.2 \\
\rowcolor{highlight} \textbf{MaskAlign} & 154M & 400K & \textbf{14.8} \\
\midrule
SiT-XL/2 & 675M & 7M & 8.3 \\
\midrule
REPA & 675M & 150K & 13.6 \\
\rowcolor{highlight} REPA + \textbf{MaskAlign}  & 728M & 150K & \textbf{10.8} \\
\midrule
REPA & 675M & 200K & 11.1 \\
ReDi & 675M & 200K & 12.5 \\
REG & 677M & 200K & 5.0 \\
\rowcolor{highlight} \textbf{MaskAlign} & 732M & 200K & \textbf{4.0} \\
\midrule
REPA & 675M & 400K & 7.9 \\
ReDi & 675M & 400K & 7.5 \\
REG & 677M & 400K & 3.4 \\
\rowcolor{highlight} \textbf{MaskAlign} & 732M & 400K & \textbf{2.8} \\
\midrule
REPA & 675M & 1M & 6.4 \\
ReDi & 675M & 1M & 5.1 \\
REG & 677M & 1M & 2.7 \\
\rowcolor{highlight} \textbf{MaskAlign} & 732M & 1M & \textbf{2.4} \\
\midrule
REG & 677M & 2.4M & 2.2 \\
\rowcolor{highlight} \textbf{MaskAlign} & 732M & 2.4M & \textbf{2.1} \\
\bottomrule
\end{tabular}
\end{minipage}
\hfill
\begin{minipage}{0.58\textwidth}
\centering
\caption{Comparison with state-of-the-art methods on \mbox{ImageNet} 256 $\times$ 256 with CFG.}
\scriptsize
\renewcommand{\arraystretch}{1.1}
\label{tab:with_cfg}
\begin{tabular}{lcccccc}
\toprule
{Model} & {Epochs} & {FID$\downarrow$} & {sFID$\downarrow$} & {IS$\uparrow$} & {Prec.$\uparrow$} & {Rec.$\uparrow$} \\
\midrule
\multicolumn{7}{l}{\textbf{\textit{Autoregressive Models}}} \\
VAR & 350 & 1.80 & - & 365.4 & 0.83 & 0.57 \\
MagViTv2 & 1080 & 1.78 & - & 319.4 & 0.83 & 0.57 \\
MAR & 800 & 1.55 & - & 303.7 & 0.81 & 0.62 \\
\midrule
\multicolumn{7}{l}{\textbf{\textit{Latent Diffusion Models}}} \\
LDM & 200 & 3.60 & - & 247.7 & 0.87 & 0.48 \\
U-ViT-H/2 & 240 & 2.29 & 5.68 & 263.9 & 0.82 & 0.57 \\
DiT-XL/2 & 1400 & 2.27 & 4.60 & 278.2 & 0.83 & 0.57 \\
MaskDiT & 1600 & 2.28 & 5.67 & 276.6 & 0.80 & 0.61 \\
SD-DiT & 480 & 3.23 & - & 270.3 & 0.82 & 0.59 \\
SiT-XL/2 & 1400 & 2.06 & 4.50 & 270.3 & 0.82 & 0.59 \\
FasterDiT & 400 & 2.03 & 4.63 & 264.0 & 0.81 & 0.60 \\
MDTV2 & 1080 & 1.58 & 4.52 & 317.7 & 0.79 & 0.65 \\
\midrule
\multicolumn{7}{l}{\textbf{\textit{Leveraging Visual Representations}}} \\
REG & 80 & 1.86 & 4.49 & 321.4 & 0.76 & \textbf{0.63} \\
\rowcolor{highlight} \textbf{MaskAlign} & 80 & \textbf{1.82} & \textbf{4.48} & \textbf{310.0} & \textbf{0.81} & \textbf{0.63} \\
\midrule
REG & 160 & 1.59 & \textbf{4.36} & \textbf{304.6} & 0.77 & \textbf{0.65} \\
\rowcolor{highlight} \textbf{MaskAlign} & 160 & \textbf{1.56} & 4.37 & 304.1 & \textbf{0.79} & \textbf{0.65} \\
\midrule
ReDi & 800 & 1.61 & 4.66 & 295.1 & \underline{0.78} & 0.64 \\
REPA & 800 & 1.42 & 4.70 & \underline{305.7} & \textbf{0.80} & 0.65 \\
REG & 800 & \underline{1.36} & \textbf{4.25} & 299.4 & 0.77 & \underline{0.66} \\
\rowcolor{highlight} \textbf{MaskAlign} & 800 & \textbf{1.35} & \underline{4.31} & \textbf{312.9} & \underline{0.78} & \textbf{0.67} \\
\bottomrule
\end{tabular}
\end{minipage}
\end{table}

\section{Experiments}
\subsection{Experimental Setup}

\paragraph{Implementation Details.}
\label{sec:implementation}
We follow the standard training procedures of SiT and REG. We conduct experiments on ImageNet, where all images are center-cropped and resized to $256 \times 256$ following the ADM preprocessing protocol. Each image is then encoded into a latent representation $z$ using the Stable Diffusion VAE. We adopt SiT-B/2 and SiT-XL/2 as the backbone architecture. For fair comparison, we use a fixed batch size of 256 and adopt the same learning rate and exponential moving average (EMA) settings as REG. More implementation details are provided in the Appendix.

\paragraph{Evaluation Protocol.}
To evaluate image generation quality from multiple aspects, we report a set of standard quantitative metrics. Specifically, we use Fréchet Inception Distance (FID)~\cite{fid} to measure sample realism, structural FID (sFID)~\cite{sfid} to evaluate spatial coherence, and Inception Score (IS)~\cite{is} to assess class-conditional diversity. We also report precision (Prec.) to measure sample fidelity and recall (Rec.)~\cite{recall} to evaluate coverage of the target distribution. All metrics are computed using 50K generated images for reliable evaluation. Following REPA, we use the SDE Euler-Maruyama solver with 250 sampling steps. Full details of the evaluation protocol are provided in the Appendix.

\paragraph{Accelerating Training Convergence.}
Table~\ref{tab:wio_cfg} reports the FID scores of different alignment-based training methods on ImageNet $256 \times 256$ without classifier-free guidance (CFG). Across different backbones and training budgets, our method consistently achieves the best FID among methods evaluated at the same number of training iterations, showing its effectiveness in accelerating training convergence.

Figure~\ref{fig:teaser} further compares the convergence curves of SiT-XL/2, SiT-XL/2 + REPA, and SiT-XL/2 + MaskAlign. To make the speedup comparison explicit, we measure the number of training iterations required to reach the same FID level. MaskAlign reaches the 8.3 FID level about $77\times$ faster than vanilla SiT-XL/2, and reaches the 5.9 FID level about $30\times$ faster than SiT-XL/2 + REPA. This shows that MaskAlign does not merely improve FID at fixed training budgets, but also reaches comparable generation quality with substantially fewer training iterations.

On SiT-B/2, our method improves REG from 15.2 to 14.8 FID at 400K iterations. On the larger SiT-XL/2 backbone, our method also brings consistent gains over REG, reducing FID from 5.0 to 4.0 at 200K iterations, from 3.4 to 2.8 at 400K iterations, and from 2.7 to 2.4 at 1M iterations. At the longer 2.4M training budget, our method further improves the FID from 2.2 to 2.1. These results indicate that MaskAlign remains effective from early training stages to longer training schedules. In addition, our method is not limited to REG. When applied to REPA, our method reduces the FID from 13.6 to 10.8 at 150K iterations, demonstrating that random token-subset alignment can also improve standard representation alignment. More experimental comparisons are provided in the Appendix.

\begin{table}[t]
\centering
\vspace{-1.0em}

\begin{minipage}[t]{0.48\linewidth}
\centering
\caption{Ablation study on token masking and token mixing. All experiments are conducted on ImageNet $256 \times 256$ using SiT-XL/2 models trained for 600K iterations without CFG.}
\label{tab:ablation_token_mixing_and_masking}
\resizebox{\linewidth}{!}{
\footnotesize
\begin{tabular}{l|ccc}
\toprule
Method & FID$\downarrow$ & sFID$\downarrow$ & IS$\uparrow$ \\
\midrule
\rowcolor{highlight}  MaskAlign & \textbf{2.67} & \textbf{4.79} & \textbf{198.10} \\
w/o Mixing & 3.54 & 6.65 & 194.51 \\
w/o Masking & 3.20 & 4.92 & 188.84 \\
w/o Both & 3.01 & 4.88 & 193.16 \\
\bottomrule
\end{tabular}
}
\end{minipage}
\hfill
\begin{minipage}[t]{0.48\linewidth}
\centering
\caption{Computational cost and performance comparison on ImageNet $256\times256$ at 400K training iterations.
Time denotes the average training time per iteration in seconds.
Both methods use the SiT-XL/2 backbone and the same GPU hardware.}
\label{tab:computation}
\resizebox{\linewidth}{!}{
\footnotesize
\renewcommand{\arraystretch}{1.2}
\begin{tabular}{l|ccc|c}
\toprule
Method & Params & Time & Tokens & FID $\downarrow$ \\
\midrule
REG & 677M & 0.359 & 257 & 3.4 \\
\rowcolor{highlight}  Ours & 732M & 0.317 & 193 & \textbf{2.8} \\
\bottomrule
\end{tabular}
}
\end{minipage}

\vspace{-1.0em}
\end{table}
\paragraph{Comparison with SOTA Methods.}
Table~\ref{tab:with_cfg} compares MaskAlign with recent generative models on ImageNet $256 \times 256$ with classifier-free guidance (CFG). MaskAlign achieves competitive performance while requiring substantially fewer training epochs than many prior diffusion-transformer baselines. At 80 epochs, MaskAlign improves REG from 1.86 to 1.82 FID and increases precision from 0.76 to 0.81, while maintaining the same recall. This model already achieves lower FID than the vanilla SiT-XL/2 trained for 1,400 epochs. At 160 epochs, MaskAlign further improves REG from 1.59 to 1.56 FID and increases precision from 0.77 to 0.79. Under the 800-epoch schedule, MaskAlign reaches 1.35 FID, slightly improving over REG and achieving higher IS and recall. These results indicate that token-subset representation alignment provides consistent gains under both short and long training schedules.

\paragraph{Computational Cost Comparison.}
Table~\ref{tab:computation} compares REG and MaskAlign at 400K training iterations using the same SiT-XL/2 backbone and GPU hardware. Although MaskAlign introduces about $8\%$ more parameters, random token masking reduces the number of input tokens from 257 to 193 and lowers the training time per step from 0.359s to 0.317s. This corresponds to a $24.9\%$ reduction in tokens and an $11.6\%$ reduction in time. Together with the faster convergence shown in Figure~\ref{fig:teaser}, this indicates that MaskAlign improves training efficiency from two aspects: it reaches the same FID level with fewer iterations and also reduces the per-step training cost. Meanwhile, MaskAlign improves FID from 3.4 to 2.8, demonstrating better sample quality with lower per-step computational cost.

\subsection{Ablation}
\paragraph{Effect of Token Masking and Token Mixing.}
We ablate the effects of pre-mask token mixing and random token masking by removing each component separately. As shown in Table~\ref{tab:ablation_token_mixing_and_masking}, the full model achieves the best performance across all metrics, indicating that both components are important for MaskAlign. Removing pre-mask token mixing leads to the worst FID and sFID, suggesting that directly applying random masking without first sharing information across tokens can severely disrupt the input token representations. Removing random masking also degrades performance, reducing the method to a token-mixing-only variant that performs worse than the baseline. These results show that token mixing and random masking are complementary: pre-mask token mixing reduces the information loss caused by dropping tokens, while random masking provides the token-subset training signal needed for more stable alignment.

\begin{wraptable}{r}{0.4\linewidth}
  \centering
  \vspace{-1.0em}
  \caption{Ablation study on the mask ratio. All models are trained for 400K iterations without CFG.}
  \label{tab:ablation_mask_ratio}
  \resizebox{\linewidth}{!}{
  \footnotesize
    \begin{tabular}{l|ccc}
    \toprule
    Mask Ratio & FID$\downarrow$ & sFID$\downarrow$ & IS$\uparrow$ \\
    \midrule
    0 & 3.52 & 4.90 & 184.13 \\
    \rowcolor{highlight} 0.25 (Ours) & \textbf{2.84} & \textbf{4.85} & \textbf{194.57} \\
    0.5 & 3.15 & 5.08 & 188.38 \\
    0.75 & 5.82 & 5.29 & 152.28 \\
    \bottomrule
    \end{tabular}
  }
  \vspace{-1.0em}
\end{wraptable}
\paragraph{Effect of Mask Ratio.}
We study the effect of the mask ratio by training models with different ratios for 400K iterations. As shown in Table~\ref{tab:ablation_mask_ratio}, a moderate mask ratio of 0.25 achieves the best performance, reducing FID from 3.52 without masking to 2.84. Increasing the mask ratio to 0.5 weakens the improvement, while an excessively high mask ratio of 0.75 severely degrades performance. These results suggest that random token masking should provide sufficient token-subset perturbations to regularize alignment, while still preserving enough input information for stable training.

\begin{wraptable}{r}{0.4\linewidth}
  \centering
  \vspace{-1.0em}
  \caption{Ablation study on the number of pre-mask token mixing layers. All models are trained for 400K iterations without CFG.}
  \label{tab:ablation_mixing_layers}
  \resizebox{\linewidth}{!}{
  \footnotesize
        \begin{tabular}{l|ccc}
        \toprule
        Mixing Layers & FID$\downarrow$ & sFID$\downarrow$ & IS$\uparrow$ \\
        \midrule
        1 & 3.23 & 4.93 & 188.49 \\
        \rowcolor{highlight} 2 (Ours) & \textbf{2.84} & \textbf{4.85} & \textbf{194.57} \\
        3 & 3.02 & 4.88 & 192.54 \\
        \bottomrule
        \end{tabular}
  }
  \vspace{-1.0em}
\end{wraptable}
\paragraph{Effect of Mixing Layers.}
We study the effect of the number of pre-mask token mixing layers. As shown in Table~\ref{tab:ablation_mixing_layers}, using two mixing layers achieves the best performance, reducing FID to 2.84. With only one mixing layer, the model obtains a higher FID of 3.23, suggesting that insufficient token mixing cannot fully compensate for the information disruption caused by random masking. 
Increasing the number of mixing layers to three also degrades performance, likely because excessive mixing alters the effective depth of the aligned representation and weakens the alignment supervision. These results indicate that a lightweight pre-mask token mixing block is sufficient.
\section{Conclusion}
\label{sec:conclusion}
In this paper, we present MaskAlign, a token-subset representation alignment method for efficient diffusion transformer training. Motivated by the mismatch between noisy diffusion features and clean-image reference representations, we analyze full-token alignment at the token level and observe a stable spatial preference among tokens with large alignment-gradient norms, suggesting that full-token alignment may encourage feature-fitting shortcuts that depend on the complete token set. To address this issue, MaskAlign applies representation alignment to randomly sampled token subsets and uses a lightweight pre-mask token mixing block to reduce the information loss caused by directly dropping tokens. Experiments on ImageNet $256 \times 256$ show that MaskAlign improves alignment stability under token-subset perturbations, accelerates training convergence, and achieves better generation quality with lower per-step computational cost. 

\paragraph{Limitations.} Despite these encouraging results, our study is mainly evaluated on ImageNet $256 \times 256$ with SiT-based backbones and pretrained DINOv2 features, and its generality to higher-resolution generation, text-to-image generation, and other teacher representations remains to be further explored. In addition, MaskAlign depends on design choices such as the mask ratio and the number of pre-mask token mixing layers, where overly aggressive masking or excessive mixing can degrade performance. Future work may investigate adaptive masking strategies and broader model families to better understand the scope and robustness of token-subset representation alignment.

%%%%%%%%%%%%%%%%%%%%%%%%%%%%%%%%%%%%%%%%%%%%%%%%%%%%%%%%%%%%

\bibliographystyle{plainnat}
\bibliography{ref}

@String{Computer = "{IEEE} Computer" }

@String{Springer = "Springer-Verlag" }

@article{balaji2023ediffi,
      title={eDiff-I: Text-to-Image Diffusion Models with an Ensemble of Expert Denoisers}, 
      author={Yogesh Balaji and Seungjun Nah and Xun Huang and Arash Vahdat and Jiaming Song and Qinsheng Zhang and Karsten Kreis and Miika Aittala and Timo Aila and Samuli Laine and Bryan Catanzaro and Tero Karras and Ming-Yu Liu},
      year={2022},
   journal={arXiv preprint arXiv:2211.01324}

}

@inproceedings{ldm,
    title={High-resolution image synthesis with latent diffusion models},
  author={Rombach, Robin and Blattmann, Andreas and Lorenz, Dominik and Esser, Patrick and Ommer, Bj{\"o}rn},
  booktitle={CVPR},
  year={2022}
}

@InProceedings{imagen,
  title={Photorealistic text-to-image diffusion models with deep language understanding},
  author={Saharia, Chitwan and Chan, William and Saxena, Saurabh and Li, Lala and Whang, Jay and Denton, Emily L and Ghasemipour, Kamyar and Gontijo Lopes, Raphael and Karagol Ayan, Burcu and Salimans, Tim and others},
  booktitle={NeurIPS},
  year={2022}
}

@article{nichol2021glide,
  title={Glide: Towards photorealistic image generation and editing with text-guided diffusion models},
  author={Nichol, Alex and Dhariwal, Prafulla and Ramesh, Aditya and Shyam, Pranav and Mishkin, Pamela and McGrew, Bob and Sutskever, Ilya and Chen, Mark},
  journal={arXiv preprint arXiv:2112.10741},
  year={2021}
}

@article{vae,
  title={Auto-encoding variational bayes},
  author={Kingma, Diederik P and Welling, Max},
  journal={arXiv preprint arXiv:1312.6114},
  year={2013}
}

@article{ddim,
  title={Denoising diffusion implicit models},
  author={Song, Jiaming and Meng, Chenlin and Ermon, Stefano},
  journal={arXiv preprint arXiv:2010.02502},
  year={2020}
}

@InProceedings{ddpm,
  title={Denoising diffusion probabilistic models},
  author={Ho, Jonathan and Jain, Ajay and Abbeel, Pieter},
  booktitle={NeurIPS},
  year={2020}
}

@InProceedings{diffusion,
      title={Deep Unsupervised Learning using Nonequilibrium Thermodynamics}, 
      author={Jascha Sohl-Dickstein and Eric A. Weiss and Niru Maheswaranathan and Surya Ganguli},
  booktitle={ICML},
  year={2015}
}

@misc{flux,
    author={Black Forest Labs},
    title={FLUX},
    year={2024},
    howpublished={\url{https://github.com/black-forest-labs/flux}},
}

@inproceedings{dit,
  title={Scalable diffusion models with transformers},
  author={Peebles, William and Xie, Saining},
  booktitle={Proceedings of the IEEE/CVF international conference on computer vision},
  pages={4195--4205},
  year={2023}
}

@inproceedings{sit,
  title={Sit: Exploring flow and diffusion-based generative models with scalable interpolant transformers},
  author={Ma, Nanye and Goldstein, Mark and Albergo, Michael S and Boffi, Nicholas M and Vanden-Eijnden, Eric and Xie, Saining},
  booktitle={European Conference on Computer Vision},
  pages={23--40},
  year={2024},
  organization={Springer}
}

@article{gao2023mdtv2,
  title={Mdtv2: Masked diffusion transformer is a strong image synthesizer},
  author={Gao, Shanghua and Zhou, Pan and Cheng, Ming-Ming and Yan, Shuicheng},
  journal={arXiv preprint arXiv:2303.14389},
  year={2023}
}

@article{maskdit,
  title={Fast training of diffusion models with masked transformers},
  author={Zheng, Hongkai and Nie, Weili and Vahdat, Arash and Anandkumar, Anima},
  journal={arXiv preprint arXiv:2306.09305},
  year={2023}
}

@inproceedings{krause2025tread,
  title={Tread: Token routing for efficient architecture-agnostic diffusion training},
  author={Krause, Felix and Phan, Timy and Gui, Ming and Baumann, Stefan Andreas and Hu, Vincent Tao and Ommer, Bj{\"o}rn},
  booktitle={Proceedings of the IEEE/CVF International Conference on Computer Vision},
  pages={15703--15713},
  year={2025}
}

@inproceedings{microdiffusion,
  title={Stretching each dollar: Diffusion training from scratch on a micro-budget},
  author={Sehwag, Vikash and Kong, Xianghao and Li, Jingtao and Spranger, Michael and Lyu, Lingjuan},
  booktitle={Proceedings of the IEEE/CVF Conference on Computer Vision and Pattern Recognition},
  pages={28596--28608},
  year={2025}
}

@article{repa,
  title={Representation alignment for generation: Training diffusion transformers is easier than you think},
  author={Yu, Sihyun and Kwak, Sangkyung and Jang, Huiwon and Jeong, Jongheon and Huang, Jonathan and Shin, Jinwoo and Xie, Saining},
  journal={arXiv preprint arXiv:2410.06940},
  year={2024}
}

@article{reg,
  title={Representation entanglement for generation: Training diffusion transformers is much easier than you think},
  author={Wu, Ge and Zhang, Shen and Shi, Ruijing and Gao, Shanghua and Chen, Zhenyuan and Wang, Lei and Chen, Zhaowei and Gao, Hongcheng and Tang, Yao and Yang, Jian and others},
  journal={arXiv preprint arXiv:2507.01467},
  year={2025}
}

@article{redi,
  title={Boosting generative image modeling via joint image-feature synthesis},
  author={Kouzelis, Theodoros and Karypidis, Efstathios and Kakogeorgiou, Ioannis and Gidaris, Spyros and Komodakis, Nikos},
  journal={arXiv preprint arXiv:2504.16064},
  year={2025}
}

@article{reglue,
  title={REGLUE Your Latents with Global and Local Semantics for Entangled Diffusion},
  author={Petsangourakis, Giorgos and Sgouropoulos, Christos and Psomas, Bill and Giannakopoulos, Theodoros and Sfikas, Giorgos and Kakogeorgiou, Ioannis},
  journal={arXiv preprint arXiv:2512.16636},
  year={2025}
}

@article{haste,
  title={REPA Works Until It Doesn't: Early-Stopped, Holistic Alignment Supercharges Diffusion Training},
  author={Wang, Ziqiao and Zhao, Wangbo and Zhou, Yuhao and Li, Zekai and Liang, Zhiyuan and Shi, Mingjia and Zhao, Xuanlei and Zhou, Pengfei and Zhang, Kaipeng and Wang, Zhangyang and others},
  journal={arXiv preprint arXiv:2505.16792},
  year={2025}
}

@article{lipman2022flow,
  title={Flow matching for generative modeling},
  author={Lipman, Yaron and Chen, Ricky TQ and Ben-Hamu, Heli and Nickel, Maximilian and Le, Matt},
  journal={arXiv preprint arXiv:2210.02747},
  year={2022}
}

@inproceedings{deng2009imagenet,
  title={Imagenet: A large-scale hierarchical image database},
  author={Deng, Jia and Dong, Wei and Socher, Richard and Li, Li-Jia and Li, Kai and Fei-Fei, Li},
  booktitle={2009 IEEE conference on computer vision and pattern recognition},
  pages={248--255},
  year={2009},
  organization={Ieee}
}

@article{baldi2013understanding,
  title={Understanding dropout},
  author={Baldi, Pierre and Sadowski, Peter J},
  journal={Advances in neural information processing systems},
  volume={26},
  year={2013}
}

@article{wager2013dropout,
  title={Dropout training as adaptive regularization},
  author={Wager, Stefan and Wang, Sida and Liang, Percy S},
  journal={Advances in neural information processing systems},
  volume={26},
  year={2013}
}

@article{fid,
  title={Gans trained by a two time-scale update rule converge to a local nash equilibrium},
  author={Heusel, Martin and Ramsauer, Hubert and Unterthiner, Thomas and Nessler, Bernhard and Hochreiter, Sepp},
  journal={Advances in neural information processing systems},
  volume={30},
  year={2017}
}

@article{sfid,
  title={Generating images with sparse representations},
  author={Nash, Charlie and Menick, Jacob and Dieleman, Sander and Battaglia, Peter W},
  journal={arXiv preprint arXiv:2103.03841},
  year={2021}
}

@article{is,
  title={Improved techniques for training gans},
  author={Salimans, Tim and Goodfellow, Ian and Zaremba, Wojciech and Cheung, Vicki and Radford, Alec and Chen, Xi},
  journal={Advances in neural information processing systems},
  volume={29},
  year={2016}
}

@article{recall,
  title={Improved precision and recall metric for assessing generative models},
  author={Kynk{\"a}{\"a}nniemi, Tuomas and Karras, Tero and Laine, Samuli and Lehtinen, Jaakko and Aila, Timo},
  journal={Advances in neural information processing systems},
  volume={32},
  year={2019}
}

@inproceedings{leng2025repae,
  title={Repa-e: Unlocking vae for end-to-end tuning of latent diffusion transformers},
  author={Leng, Xingjian and Singh, Jaskirat and Hou, Yunzhong and Xing, Zhenchang and Xie, Saining and Zheng, Liang},
  booktitle={Proceedings of the IEEE/CVF International Conference on Computer Vision},
  pages={18262--18272},
  year={2025}
}

@article{singh2025matters,
  title={What matters for Representation Alignment: Global Information or Spatial Structure?},
  author={Singh, Jaskirat and Leng, Xingjian and Wu, Zongze and Zheng, Liang and Zhang, Richard and Shechtman, Eli and Xie, Saining},
  journal={arXiv preprint arXiv:2512.10794},
  year={2025}
}

%%%%%%%%%%%%%%%%%%%%%%%%%%%%%%%%%%%%%%%%%%%%%%%%%%%%%%%%%%%%

\newpage
\appendix

\section{Experimental Setup}
Table~\ref{tab:appendix_hyperparams} summarizes the hyperparameter settings of MaskAlign for SiT-B/2 and SiT-XL/2. Following the experimental protocol of REPA, we train models in the latent space with v-prediction and use the Euler-Maruyama solver with 250 sampling steps for evaluation. Across both model scales, we use DINOv2-B as the pretrained vision encoder, cosine similarity for representation alignment, two pre-mask token mixing layers, and a mask ratio of $25\%$. The alignment weight is set to $\lambda=0.5$, and the class-token prediction weight is set to $\beta=0.03$. For optimization, we use AdamW with a batch size of 256 and a learning rate of $1\times10^{-4}$.
\begin{table}[ht]
\centering
\caption{Hyperparameter settings across different model scales.}
\label{tab:appendix_hyperparams}
\begin{tabular}{lcc}
\toprule
Backbone & SiT-B & SiT-XL \\
\midrule
\multicolumn{3}{l}{\textbf{Architecture}} \\
\#Params & 154M & 732M \\
Input & $32 \times 32 \times 4$ & $32 \times 32 \times 4$ \\
Layers & 12 & 28 \\
Hidden dim. & 768 & 1,152 \\
Num. heads & 12 & 16 \\
\midrule
\multicolumn{3}{l}{\textbf{MaskAlign settings}} \\
$\beta$ & 0.03 & 0.03 \\
$\lambda$ & 0.5 & 0.5 \\
Alignment depth & 4 & 8 \\
Mixing Layers & 2 & 2 \\
Mask Ratio & 25\% & 25\% \\
$\text{sim}(\cdot, \cdot)$ & cos. sim. & cos. sim. \\
Encoder $\mathcal{E}_{VF}(I)$ & DINOv2-B & DINOv2-B \\
\midrule
\multicolumn{3}{l}{\textbf{Optimization}} \\
Batch size & 256 & 256 \\
Optimizer & AdamW & AdamW \\
lr & 0.0001 & 0.0001 \\
$(\beta_1, \beta_2)$ & (0.9, 0.999) & (0.9, 0.999) \\
\midrule
\multicolumn{3}{l}{\textbf{Interpolants}} \\
$\alpha_t$ & $1 - t$ & $1 - t$ \\
$\sigma_t$ & $t$ & $t$ \\
$w_t$ & $\sigma_t$ & $\sigma_t$ \\
Training objective & v-prediction & v-prediction \\
Sampler & Euler-Maruyama & Euler-Maruyama \\
Sampling steps & 250 & 250 \\
\bottomrule
\end{tabular}
\end{table}

\section{Additional Token-Level Alignment Heatmaps}
Table~\ref{tab:appendix_heatmaps} provides additional heatmaps of the token-level alignment-gradient distribution under different timesteps and training iterations. Each heatmap shows the probability that each spatial position appears among the top-$10\%$ tokens ranked by alignment-gradient norm. Across different timesteps and checkpoints, the high-gradient tokens exhibit non-uniform spatial patterns, further supporting our observation that full-token representation alignment does not affect all patch tokens uniformly.
\begin{table}[htbp]
\centering
\caption{
Additional alignment-gradient heatmaps across timesteps and training iterations.
Rows denote training iterations, and columns denote timesteps.
Each heatmap shows the spatial probability of top-$10\%$ alignment-gradient tokens, using the same color range $[0,0.8]$.
}
\label{tab:appendix_heatmaps}

\setlength{\tabcolsep}{2pt} % 控制列间距，默认通常是 6pt
\renewcommand{\arraystretch}{0.9} % 控制行间距，可按需调整

\begin{tabular}{c@{\hspace{2pt}}c@{\hspace{2pt}}c@{\hspace{2pt}}c@{\hspace{2pt}}c@{\hspace{2pt}}c}
\quad & \textbf{t=0.1} & \textbf{t=0.3} & \textbf{t=0.5} & \textbf{t=0.7} & \textbf{t=0.9} \\

\raisebox{1cm}{\textbf{100K}} & 
\includegraphics[width=0.17\linewidth]{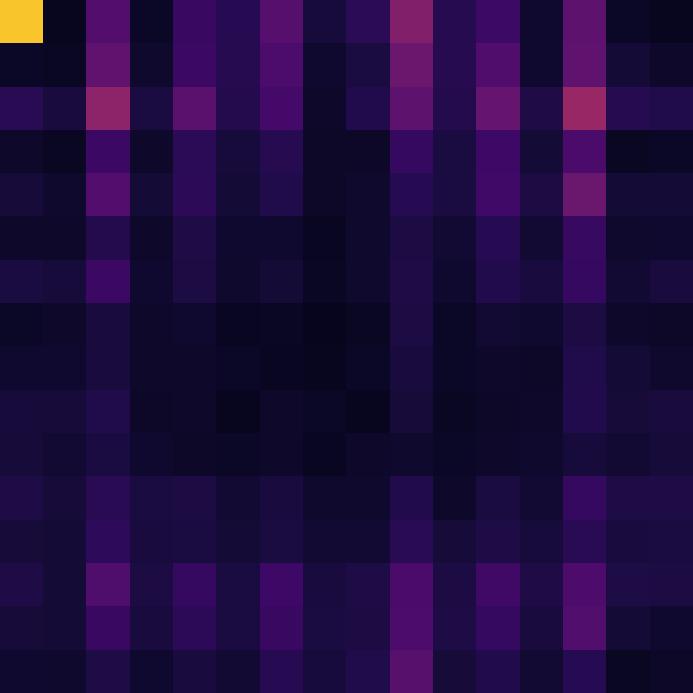} & 
\includegraphics[width=0.17\linewidth]{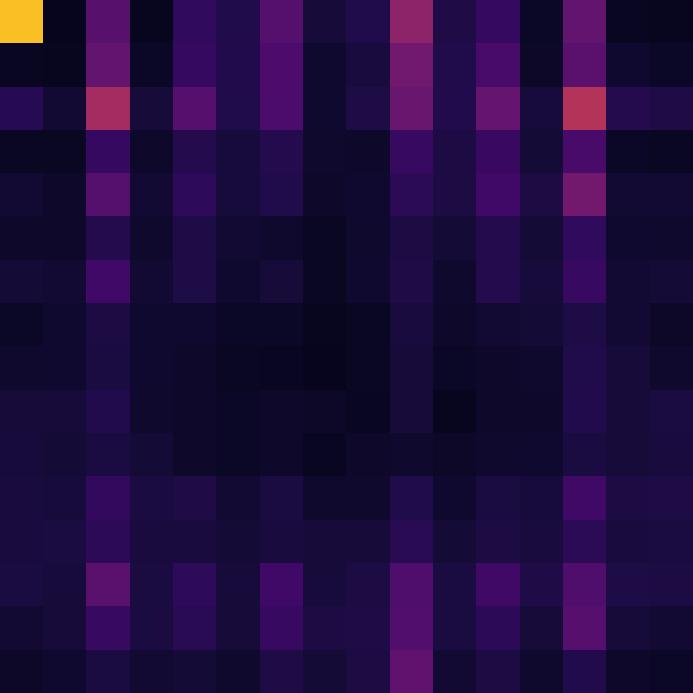} & 
\includegraphics[width=0.17\linewidth]{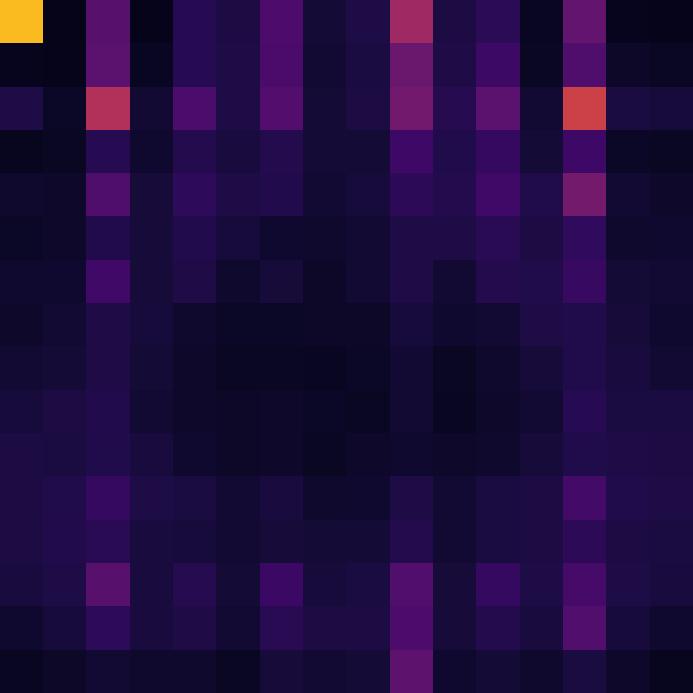} & 
\includegraphics[width=0.17\linewidth]{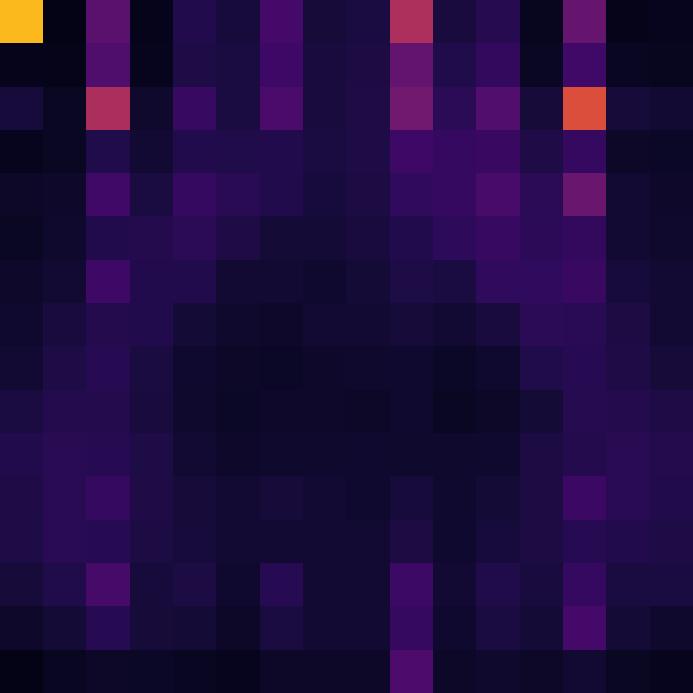} & 
\includegraphics[width=0.17\linewidth]{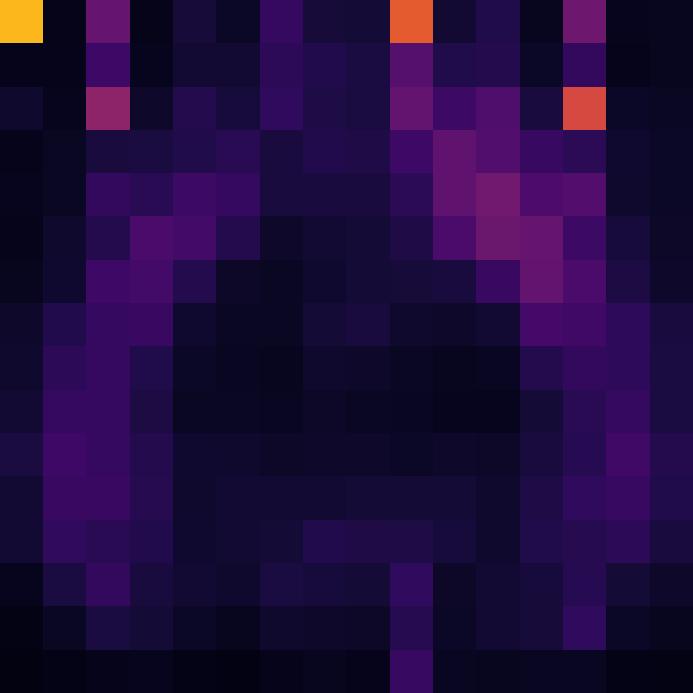} \\

\raisebox{1cm}{\textbf{500K}} & 
\includegraphics[width=0.17\linewidth]{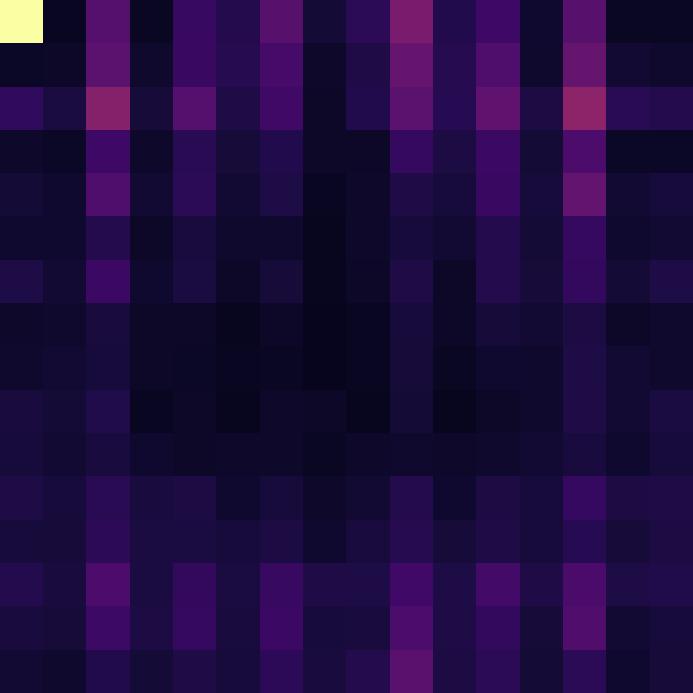} & 
\includegraphics[width=0.17\linewidth]{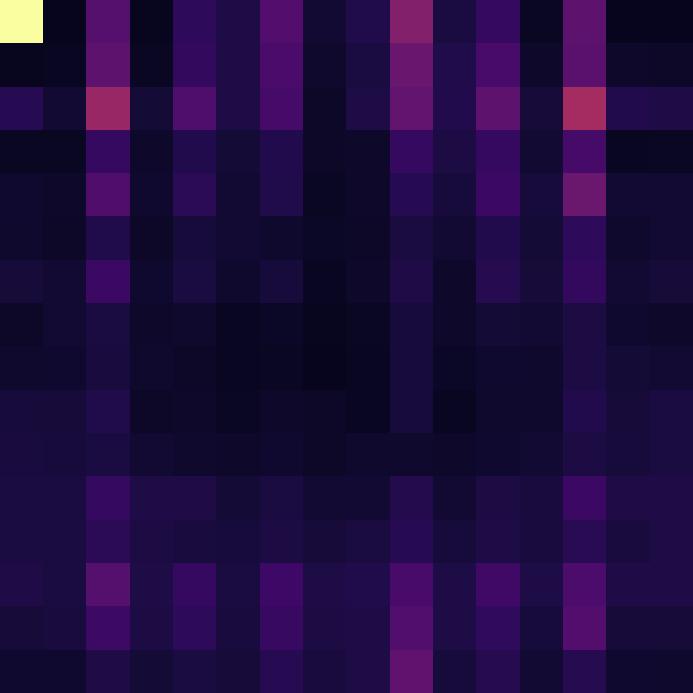} & 
\includegraphics[width=0.17\linewidth]{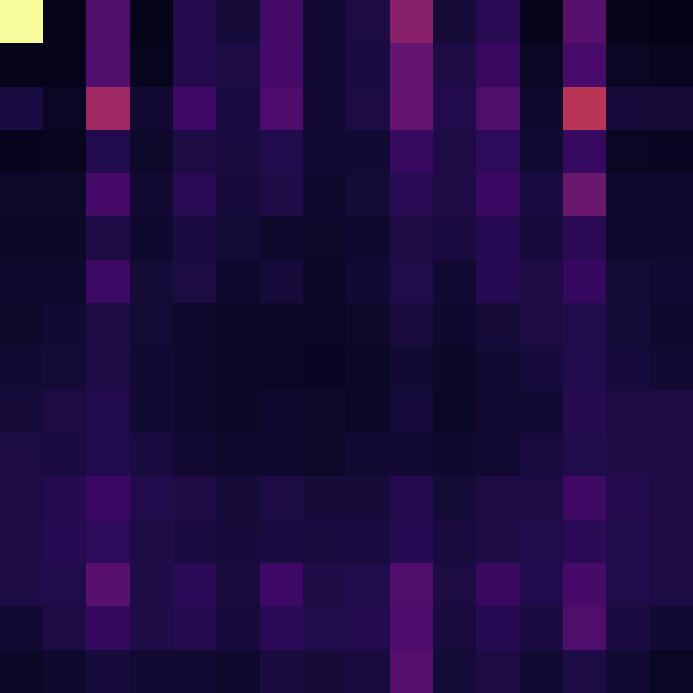} & 
\includegraphics[width=0.17\linewidth]{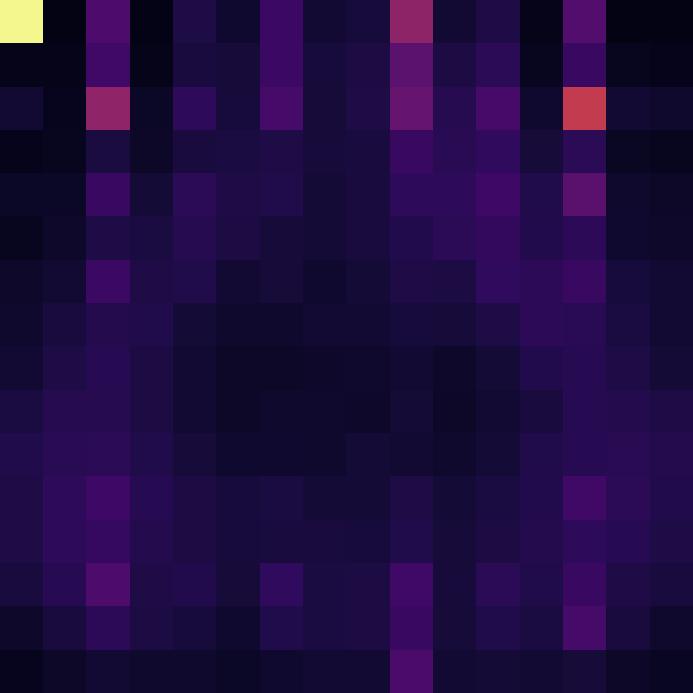} & 
\includegraphics[width=0.17\linewidth]{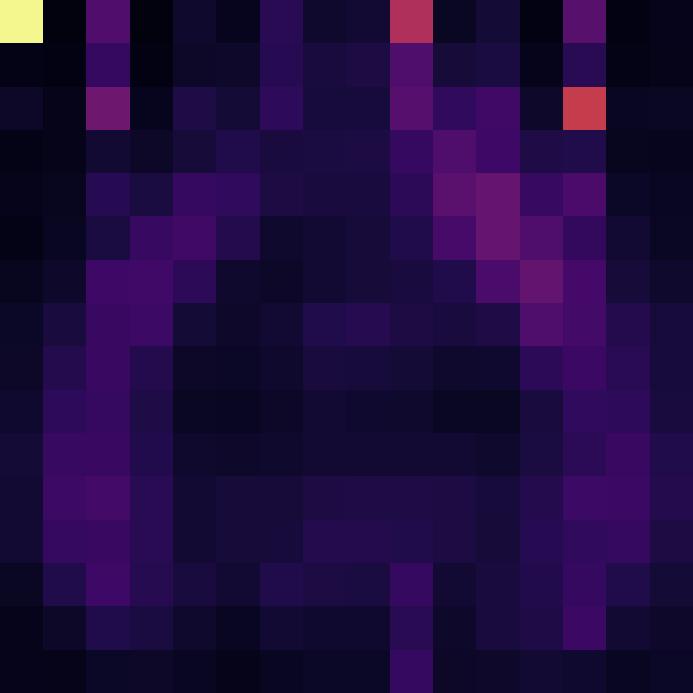} \\

\raisebox{1cm}{\textbf{1M}} & 
\includegraphics[width=0.17\linewidth]{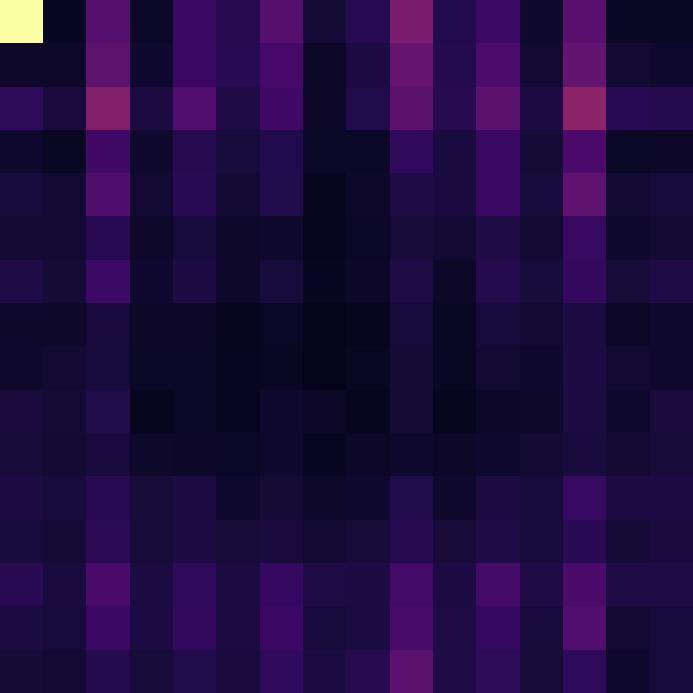} & 
\includegraphics[width=0.17\linewidth]{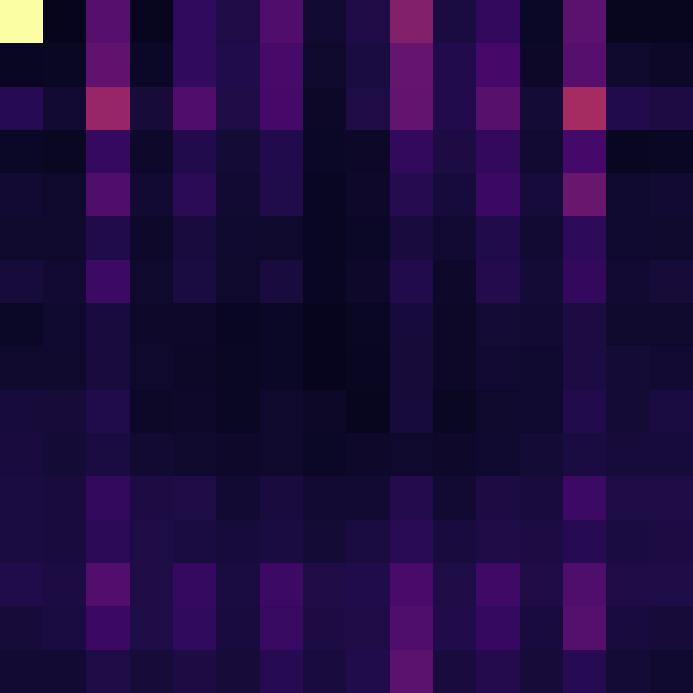} & 
\includegraphics[width=0.17\linewidth]{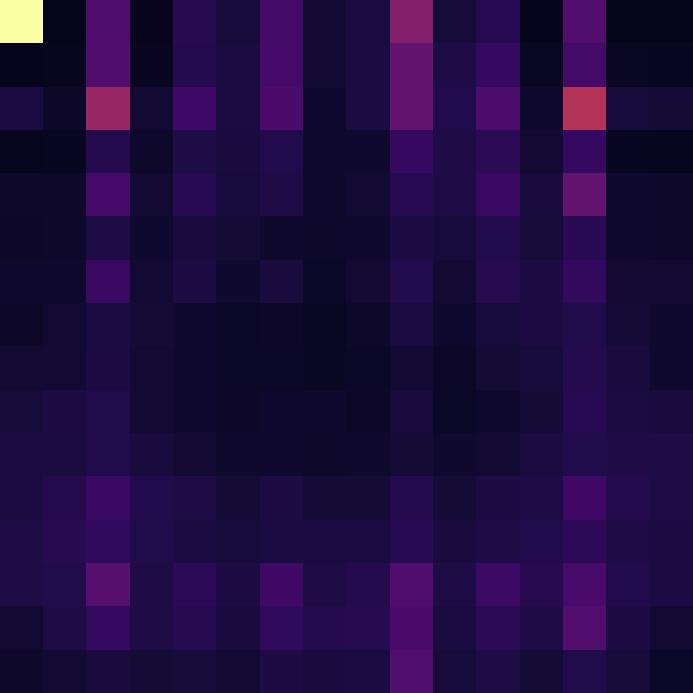} & 
\includegraphics[width=0.17\linewidth]{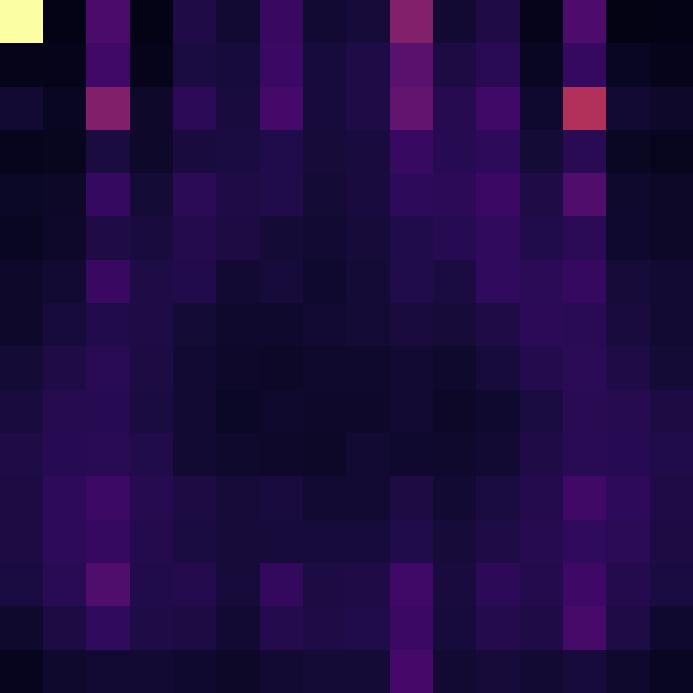} & 
\includegraphics[width=0.17\linewidth]{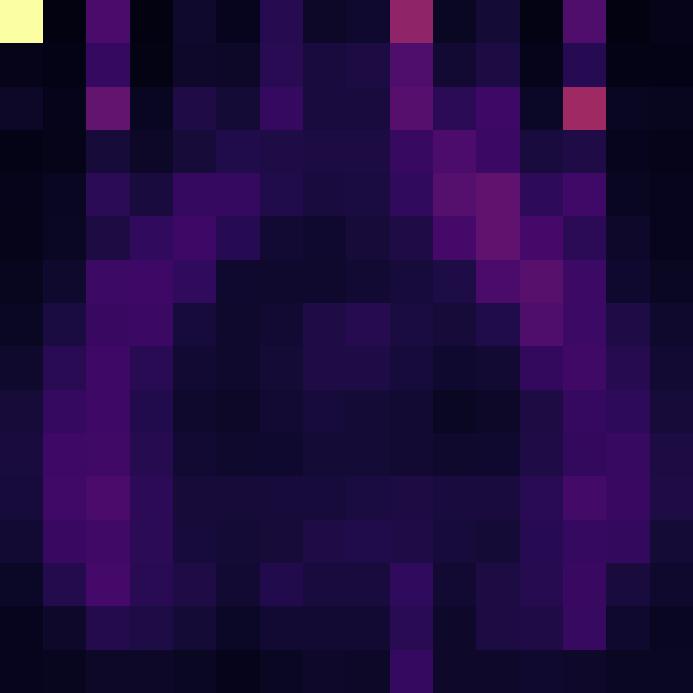} \\

\end{tabular}
\end{table}

\section{Additional Results on ImageNet}
Table~\ref{tab:appendix_results} reports additional quantitative results of MaskAlign on ImageNet $256 \times 256$ without classifier-free guidance (CFG). We evaluate MaskAlign at different training iterations to provide a more detailed view of its convergence behavior. As training proceeds, MaskAlign consistently improves generation quality, reducing FID from 22.36 at 50K iterations to 2.38 at 1M iterations. Compared with REG trained for 1M iterations, MaskAlign achieves better FID, sFID, and IS at the same training budget, while maintaining comparable precision and recall.
\begin{table}[h]
\centering
\caption{
Additional quantitative results of MaskAlign on ImageNet $256 \times 256$ without classifier-free guidance (CFG).
We report FID, sFID, IS, precision, and recall across different training iterations.
}
\label{tab:appendix_results}
\begin{tabular}{lccccccc}
\toprule
Model & \#Params & Iter. & FID$\downarrow$ & sFID$\downarrow$ & IS$\uparrow$ & Prec.$\uparrow$ & Rec.$\uparrow$ \\
\midrule
SiT-XL/2 & 675M & 7M & 8.3 & 6.32 & 131.7 & 0.68 & \textbf{0.67} \\
REG & 677M & 1M & 2.7 & 4.93 & 201.8 & 0.76 & 0.66 \\
\midrule
MaskAlign & 732M & 50K & 22.36 & 20.62 & 70.24 & 0.63 & 0.53 \\
MaskAlign & 732M & 110K & 6.34 & 6.47 & 145.79 & 0.75 & 0.58 \\
MaskAlign & 732M & 200K & 3.98 & 5.18 & 172.60 & 0.77 & 0.60 \\
MaskAlign & 732M & 400K & 2.84 & 4.85 & 194.57 & 0.77 & 0.62 \\
MaskAlign & 732M & 600K & 2.67 & 4.78 & 198.01 & \textbf{0.77} & 0.64 \\
MaskAlign & 732M & 1M & \textbf{2.38} & \textbf{4.78} & \textbf{205.37} & 0.76 & 0.65 \\

\bottomrule
\end{tabular}
\end{table}

\section{Broader Impacts}
\label{sec:broader_impacts}
This work aims to improve the efficiency of diffusion transformer training. Its potential positive impacts include reducing the computational cost of training high-quality generative models and making research on diffusion models more accessible. However, more efficient training may also lower the barrier to building image generation systems, which could increase risks such as misleading synthetic content, impersonation, and biases inherited from training data or pretrained vision models. Our work does not introduce a deployed system or a new dataset, but responsible use of trained models should consider safeguards such as data curation, provenance tracking, watermarking, and controlled release when appropriate.

\section{Assets and Licenses}
\label{sec:assets_licenses}
We use ImageNet for non-commercial research and educational purposes, following its terms of access, and cite the original ImageNet paper. We use the Stable Diffusion VAE released by Stability AI under the MIT License to encode images into latent representations. We use DINOv2-B as the pretrained vision encoder for representation alignment; DINOv2 code and model weights are released under the Apache License 2.0. Our implementation also builds on the SiT, REPA, and REG codebases, which are released under the MIT License. We properly credit these prior works through citations and use the corresponding assets only for research purposes and in accordance with their licenses and terms of use.

\section{More Visualization Results}
\begin{figure}[h]
    \centering
    \setlength{\tabcolsep}{0pt}
    \renewcommand{\arraystretch}{0}

    \begin{tabular}{@{}cccccccc@{}}
        \multicolumn{2}{@{}c@{}}{\includegraphics[width=0.25\linewidth]{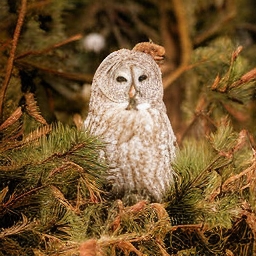}} &
        \multicolumn{2}{@{}c@{}}{\includegraphics[width=0.25\linewidth]{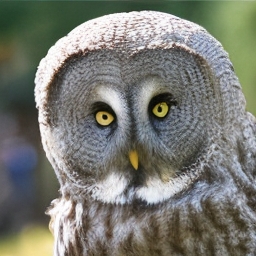}} &
        \multicolumn{2}{@{}c@{}}{\includegraphics[width=0.25\linewidth]{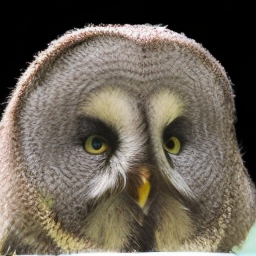}} &
        \multicolumn{2}{@{}c@{}}{\includegraphics[width=0.25\linewidth]{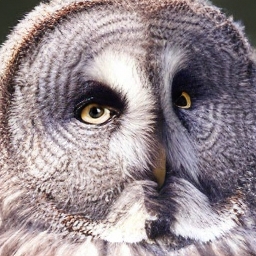}} \\
        \includegraphics[width=0.125\linewidth]{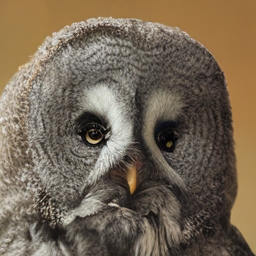} &
        \includegraphics[width=0.125\linewidth]{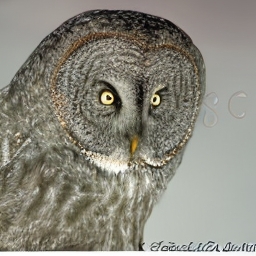} &
        \includegraphics[width=0.125\linewidth]{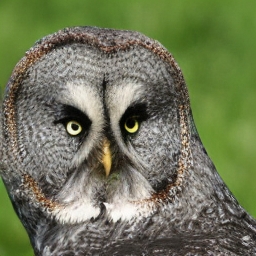} &
        \includegraphics[width=0.125\linewidth]{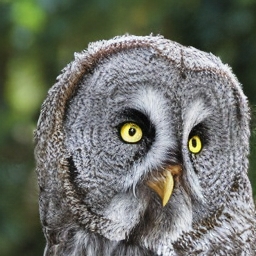} &
        \includegraphics[width=0.125\linewidth]{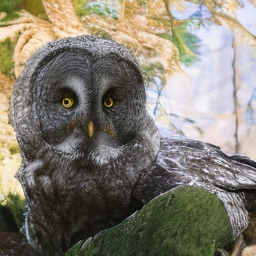} &
        \includegraphics[width=0.125\linewidth]{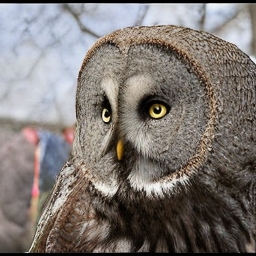} &
        \includegraphics[width=0.125\linewidth]{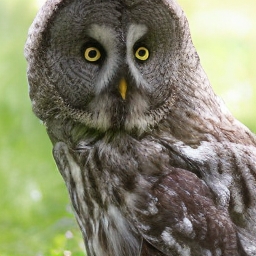} &
        \includegraphics[width=0.125\linewidth]{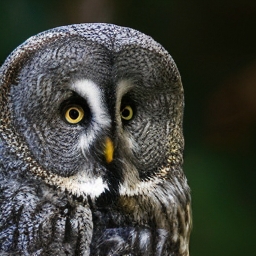}
    \end{tabular}

    \caption{Generated samples from SiT-XL/2 + MaskAlign. The class label is “great grey owl” (24).}
    \label{fig:appendix_visualization_24}
\end{figure}

\begin{figure}[h]
    \centering
    \setlength{\tabcolsep}{0pt}
    \renewcommand{\arraystretch}{0}

    \begin{tabular}{@{}cccccccc@{}}
        \multicolumn{2}{@{}c@{}}{\includegraphics[width=0.25\linewidth]{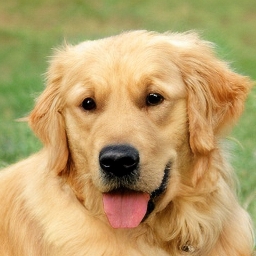}} &
        \multicolumn{2}{@{}c@{}}{\includegraphics[width=0.25\linewidth]{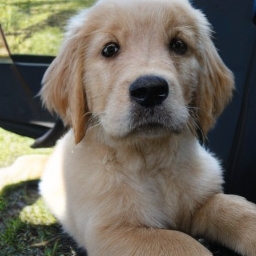}} &
        \multicolumn{2}{@{}c@{}}{\includegraphics[width=0.25\linewidth]{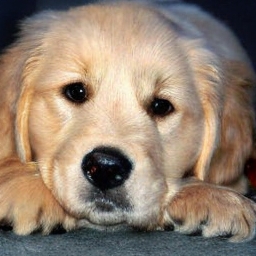}} &
        \multicolumn{2}{@{}c@{}}{\includegraphics[width=0.25\linewidth]{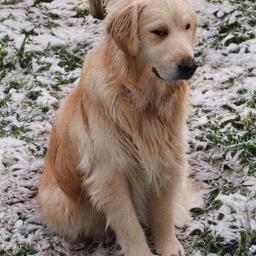}} \\
        \includegraphics[width=0.125\linewidth]{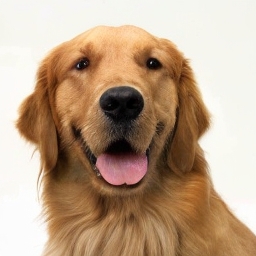} &
        \includegraphics[width=0.125\linewidth]{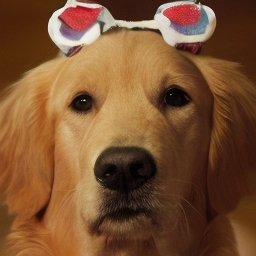} &
        \includegraphics[width=0.125\linewidth]{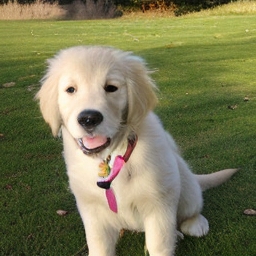} &
        \includegraphics[width=0.125\linewidth]{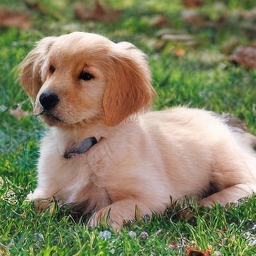} &
        \includegraphics[width=0.125\linewidth]{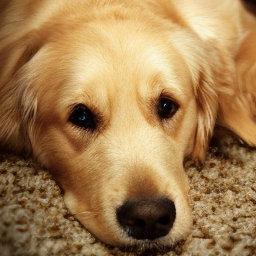} &
        \includegraphics[width=0.125\linewidth]{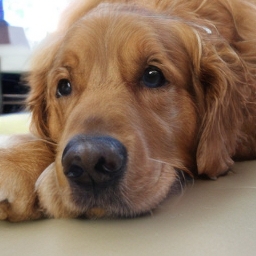} &
        \includegraphics[width=0.125\linewidth]{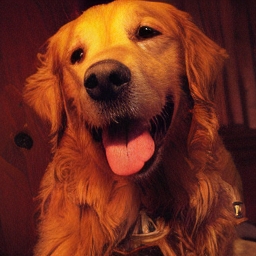} &
        \includegraphics[width=0.125\linewidth]{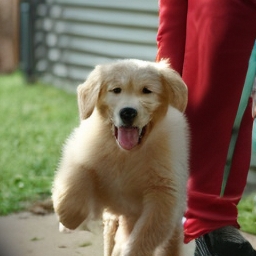}
    \end{tabular}

    \caption{Generated samples from SiT-XL/2 + MaskAlign. The class label is “golden retriever” (207).}
    \label{fig:appendix_visualization_207}
\end{figure}

\begin{figure}[h]
    \centering
    \setlength{\tabcolsep}{0pt}
    \renewcommand{\arraystretch}{0}

    \begin{tabular}{@{}cccccccc@{}}
        \multicolumn{2}{@{}c@{}}{\includegraphics[width=0.25\linewidth]{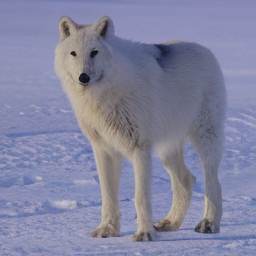}} &
        \multicolumn{2}{@{}c@{}}{\includegraphics[width=0.25\linewidth]{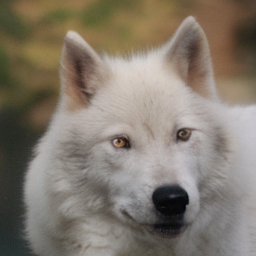}} &
        \multicolumn{2}{@{}c@{}}{\includegraphics[width=0.25\linewidth]{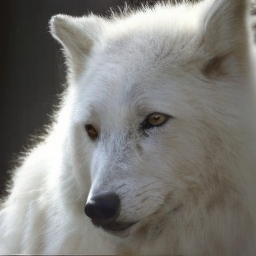}} &
        \multicolumn{2}{@{}c@{}}{\includegraphics[width=0.25\linewidth]{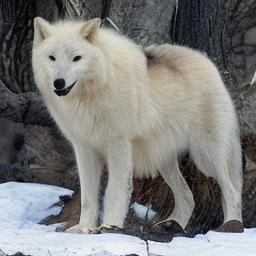}} \\
        \includegraphics[width=0.125\linewidth]{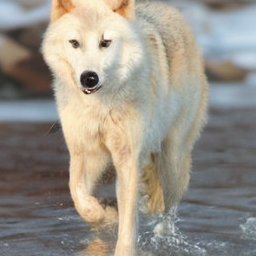} &
        \includegraphics[width=0.125\linewidth]{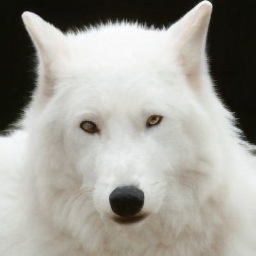} &
        \includegraphics[width=0.125\linewidth]{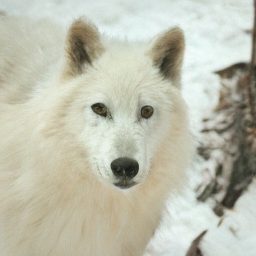} &
        \includegraphics[width=0.125\linewidth]{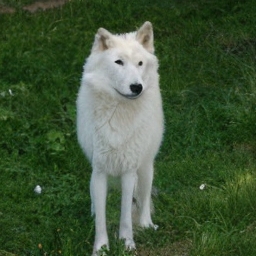} &
        \includegraphics[width=0.125\linewidth]{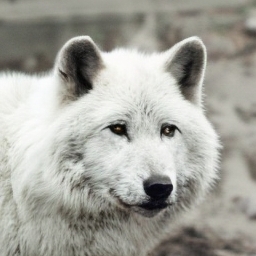} &
        \includegraphics[width=0.125\linewidth]{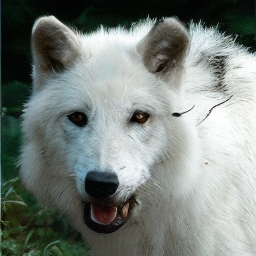} &
        \includegraphics[width=0.125\linewidth]{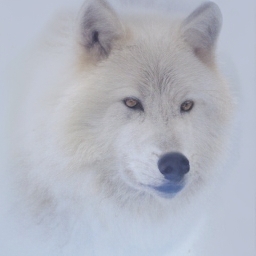} &
        \includegraphics[width=0.125\linewidth]{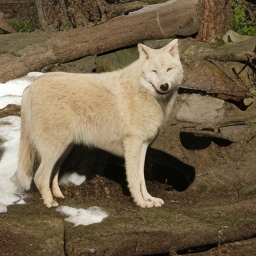}
    \end{tabular}

    \caption{Generated samples from SiT-XL/2 + MaskAlign. The class label is “arctic wolf” (270).}
    \label{fig:appendix_visualization_270}
\end{figure}

\begin{figure}[h]
    \centering
    \setlength{\tabcolsep}{0pt}
    \renewcommand{\arraystretch}{0}

    \begin{tabular}{@{}cccccccc@{}}
        \multicolumn{2}{@{}c@{}}{\includegraphics[width=0.25\linewidth]{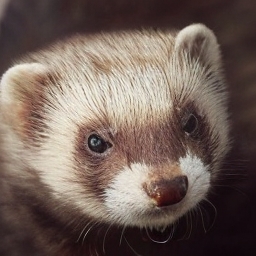}} &
        \multicolumn{2}{@{}c@{}}{\includegraphics[width=0.25\linewidth]{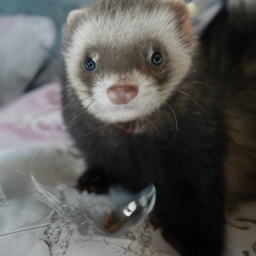}} &
        \multicolumn{2}{@{}c@{}}{\includegraphics[width=0.25\linewidth]{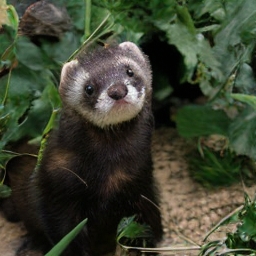}} &
        \multicolumn{2}{@{}c@{}}{\includegraphics[width=0.25\linewidth]{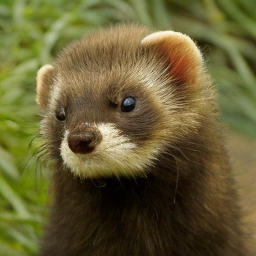}} \\
        \includegraphics[width=0.125\linewidth]{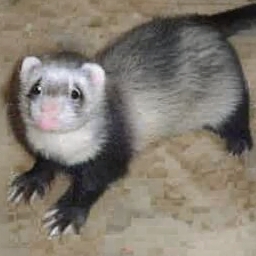} &
        \includegraphics[width=0.125\linewidth]{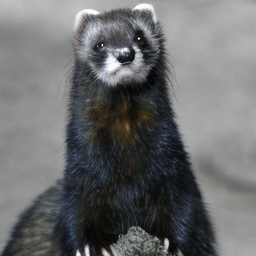} &
        \includegraphics[width=0.125\linewidth]{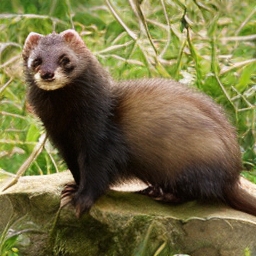} &
        \includegraphics[width=0.125\linewidth]{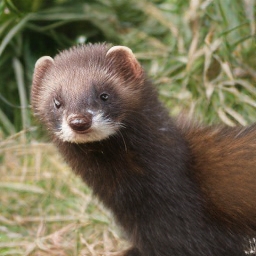} &
        \includegraphics[width=0.125\linewidth]{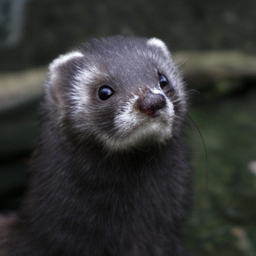} &
        \includegraphics[width=0.125\linewidth]{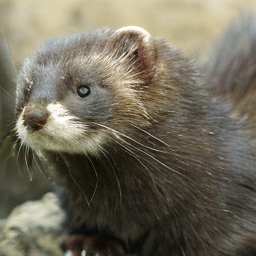} &
        \includegraphics[width=0.125\linewidth]{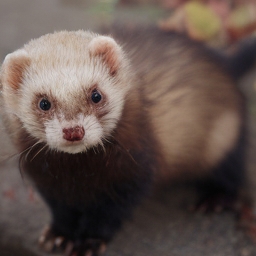} &
        \includegraphics[width=0.125\linewidth]{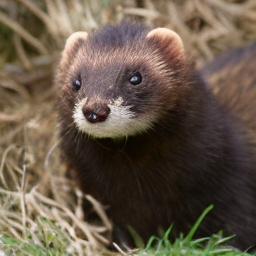}
    \end{tabular}

    \caption{Generated samples from SiT-XL/2 + MaskAlign. The class label is “polecat” (358).}
    \label{fig:appendix_visualization_358}
\end{figure}

\begin{figure}[h]
    \centering
    \setlength{\tabcolsep}{0pt}
    \renewcommand{\arraystretch}{0}

    \begin{tabular}{@{}cccccccc@{}}
        \multicolumn{2}{@{}c@{}}{\includegraphics[width=0.25\linewidth]{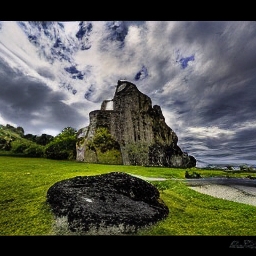}} &
        \multicolumn{2}{@{}c@{}}{\includegraphics[width=0.25\linewidth]{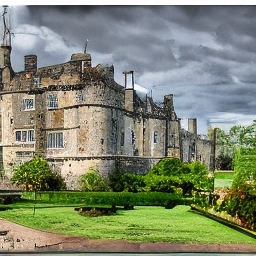}} &
        \multicolumn{2}{@{}c@{}}{\includegraphics[width=0.25\linewidth]{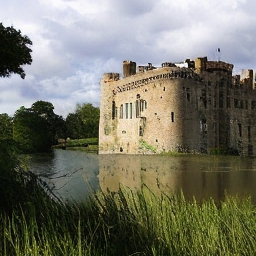}} &
        \multicolumn{2}{@{}c@{}}{\includegraphics[width=0.25\linewidth]{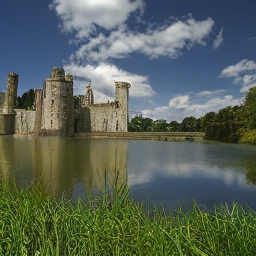}} \\
        \includegraphics[width=0.125\linewidth]{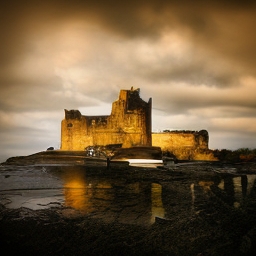} &
        \includegraphics[width=0.125\linewidth]{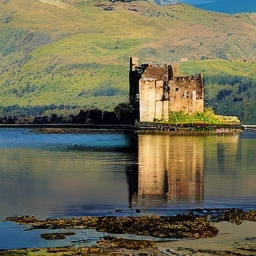} &
        \includegraphics[width=0.125\linewidth]{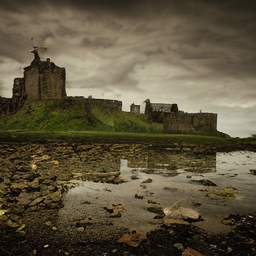} &
        \includegraphics[width=0.125\linewidth]{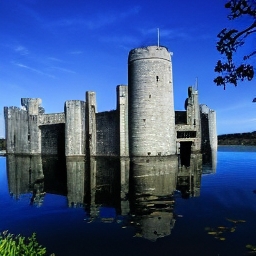} &
        \includegraphics[width=0.125\linewidth]{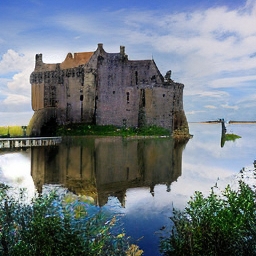} &
        \includegraphics[width=0.125\linewidth]{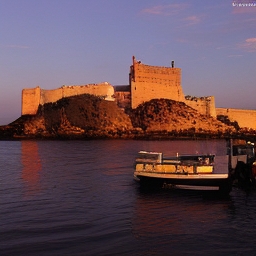} &
        \includegraphics[width=0.125\linewidth]{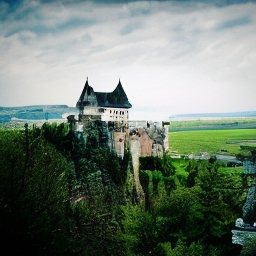} &
        \includegraphics[width=0.125\linewidth]{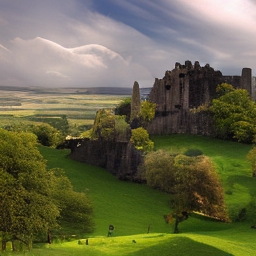}
    \end{tabular}

    \caption{Generated samples from SiT-XL/2 + MaskAlign. The class label is “castle” (483).}
    \label{fig:appendix_visualization_483}
\end{figure}

We present more visualization results of MaskAlign in Figures~\ref{fig:appendix_visualization_24}--\ref{fig:appendix_visualization_483}.


\end{document}